# A Bilinear Programming Approach for Multiagent Planning

**Marek Petrik**                                                      PETRIK@CS.UMASS.EDU

**Shlomo Zilberstein**                                               SHLOMO@CS.UMASS.EDU
*Department of Computer Science*
*University of Massachusetts, Amherst, MA 01003, USA*

## Abstract

Multiagent planning and coordination problems are common and known to be computationally hard. We show that a wide range of two-agent problems can be formulated as bilinear programs. We present a successive approximation algorithm that significantly outperforms the coverage set algorithm, which is the state-of-the-art method for this class of multiagent problems. Because the algorithm is formulated for bilinear programs, it is more general and simpler to implement. The new algorithm can be terminated at any time and–unlike the coverage set algorithm–it facilitates the derivation of a useful online performance bound. It is also much more efficient, on average reducing the computation time of the optimal solution by about four orders of magnitude. Finally, we introduce an automatic dimensionality reduction method that improves the effectiveness of the algorithm, extending its applicability to new domains and providing a new way to analyze a subclass of bilinear programs.

## 1. Introduction

We present a new approach for solving a range of multiagent planning and coordination problems using bilinear programming. The problems we focus on represent various extensions of the Markov decision process (MDP) to multiagent settings. The success of MDP algorithms for planning and learning under uncertainty has motivated researchers to extend the model to cooperative multiagent problems. One possibility is to assume that all the agents share all the information about the underlying state. This results in a multiagent Markov decision process (Boutilier, 1999), which is essentially an MDP with a factored action set. A more complex alternative is to allow only partial sharing of information among agents. In these settings, several agents–each having different partial information about the world–must cooperate with each other in order to achieve some joint objective. Such problems are common in practice and can be modeled as decentralized partially observable MDPs (DEC-POMDPs) (Bernstein, Zilberstein, & Immerman, 2000). Some refinements of this model have been studied, for example by making certain independence assumptions (Becker, Zilberstein, & Lesser, 2003) or by adding explicit communication actions (Goldman & Zilberstein, 2008). DEC-POMDPs are closely related to extensive games (Rubinstein, 1997). In fact, any DEC-POMDP represents an exponentially larger extensive game with a common objective. Unfortunately, DEC-POMDPs with just two agents are intractable in general, unlike MDPs that can be solved in polynomial time.

Despite recent progress in solving DEC-POMDPs, even state-of-the-art algorithms are generally limited to very small problems (Seuken & Zilberstein, 2008). This has motivated the development of algorithms that either solve a restricted class of problems (Becker,

Lesser, & Zilberstein, 2004; Kim, Nair, Varakantham, Tambe, & Yokoo, 2006) or provide only approximate solutions (Emery-Montemerlo, Gordon, Schneider, & Thrun, 2004; Nair, Roth, Yokoo, & Tambe, 2004; Seuken & Zilberstein, 2007). In this paper, we introduce an efficient algorithm for several restricted classes, most notably decentralized MDPs with transition and observation independence (Becker et al., 2003). For the sake of simplicity, we denote this model as DEC-MDP, although this is usually used to denote the model without the independence assumptions. The objective in these problems is to maximize the cumulative reward of a set of cooperative agents over some finite horizon. Each agent can be viewed as a single decision-maker operating on its own "local" MDP. What complicates the problem is the fact that all these MDPs are linked through a common reward function that depends on their states.

The coverage set algorithm (CSA) was the first optimal algorithm to solve efficiently transition and observation independent DEC-MDPs (Becker, Zilberstein, Lesser, & Goldman, 2004). By exploiting the fact that the interaction between the agents is limited compared to their individual local problems, CSA can solve problems that cannot be solved by the more general exact DEC-POMDP algorithms. It also exhibits good anytime behavior. However, the anytime behavior is of limited applicability because solution quality is only known in hindsight, after the algorithm terminates.

We develop a new approach to solve DEC-MDPs–as well as a range of other multiagent planning problems–by representing them as bilinear programs. We also present an efficient new algorithm for solving these kinds of separable bilinear problems. When the algorithm is applied to DEC-MDPs, it improves efficiency by several orders of magnitude compared with previous state-of-the art algorithms (Becker, 2006; Petrik & Zilberstein, 2007a). In addition, the algorithm provides useful runtime bounds on the approximation error, which makes it more useful as an anytime algorithm. Finally, the algorithm is formulated for general separable bilinear programs and therefore it can be easily applied to a range of other problems.

The rest of the paper is organized as follows. First, in Section 2, we describe the basic bilinear program formulation and how a range of multiagent planning problems can be expressed within this framework. In Section 3, we describe a new successive approximation algorithm for bilinear programs. The performance of the algorithm depends heavily on the number of interactions between the agents. To address that, we propose in Section 4 a method that automatically reduces the number of interactions and provides a bound on the degradation in solution quality. Furthermore, to be able to project the computational effort required to solve a given problem instance, we develop offline approximation bounds in Section 5. In Section 6, we examine the performance of the approach on a standard benchmark problem. We conclude with a summary of the results and a discussion of future work that could further improve the performance of this approach.

## 2. Formulating Multiagent Planning Problems as Bilinear Programs

We begin with a formal description of bilinear programs and the different types of multiagent planning problems that can be formulated as such. In addition to multiagent planning problems, bilinear programs can be used to solve a variety of other problems such as robotic manipulation (Pang, Trinkle, & Lo, 1996), bilinear separation (Bennett & Mangasarian, 1992), and even general linear complementarity problems (Mangasarian, 1995). We focus on multiagent planning problems where this formulation turns out to be particularly effective.

**Definition 1.** A *separable* bilinear program in the normal form is defined as follows:

$$
\begin{aligned}
\underset{w,x,y,z}{\text{maximize}} \quad & f(w,x,y,z) = s_1^\mathsf{T} w + r_1^\mathsf{T} x + x^\mathsf{T} C y + r_2^\mathsf{T} y + s_2^\mathsf{T} z \\
\text{subject to} \quad & A_1 x + B_1 w = b_1 \\
& A_2 y + B_2 z = b_2 \\
& w, x, y, z \geq \mathbf{0}
\end{aligned}
\tag{1}
$$

The *size* of the program is the total number of variables in $w, x, y$ and $z$. The number of variables in $y$ determines the *dimensionality* of the program[1].

Unless otherwise specified, all vectors are column vectors. We use boldface $\mathbf{0}$ and $\mathbf{1}$ to denote vectors of zeros and ones respectively of the appropriate dimensions. This program specifies two linear programs that are connected only through the nonlinear objective function term $x^\mathsf{T} C y$. The program contains two types of variables. The first type includes the variables $x, y$ that appear in the bilinear term of the objective function. The second type includes the additional variables $w, z$ that do not appear in the bilinear term. As we show later, this distinction is important because the complexity of the algorithm we propose depends mostly on the dimensionality of the problem, which is the number of variables $y$ involved in the bilinear term.

The bilinear program in Eq. (1) is *separable* because the constraints on $x$ and $w$ are independent of the constraints on $y$ and $z$. That is, the variables that participate in the bilinear term of the objective function are *independently* constrained. The theory of non-separable bilinear programs is much more complicated and the corresponding algorithms are not as efficient (Horst & Tuy, 1996). Thus, we limit the discussion in this paper to separable bilinear programs and often omit the term "separable". As discussed later in more detail, a *separable* bilinear program may be seen as a concave minimization problem with multiple local minima. It can be shown that solving this problem is NP-complete, compared to polynomial time complexity of linear programs.

In addition to the formulation of the bilinear program shown in Eq. (1), we also use the following formulation, stated in terms of inequalities:

$$
\begin{aligned}
\underset{x,y}{\text{maximize}} \quad & x^\mathsf{T} C y \\
\text{subject to} \quad & A_1 x \leq b_1 \quad x \geq \mathbf{0} \\
& A_2 y \leq b_2 \quad y \geq \mathbf{0}
\end{aligned}
\tag{2}
$$

The latter formulation can be easily transformed into the normal form using standard transformations of linear programs (Vanderbei, 2001). In particular, we can introduce slack

---

1. It is possible to define the dimensionality in terms of $x$, or the minimum of dimensions of $x$ and $y$. The issue is discussed in Appendix B.

variables $w, z$ to obtain the following identical bilinear program in the normal form:

$$
\begin{aligned}
\underset{w,x,y,z}{\text{maximize}} \quad & x^\mathsf{T} C y \\
\text{subject to} \quad & A_1 x - w = b_1 \\
& A_2 y - z = b_2 \\
& w, x, y, z \geq \mathbf{0}
\end{aligned}
\tag{3}
$$

We use the following matrix and block matrix notation in the paper. Matrices are denoted by square brackets, with columns separated by commas and rows separated by semicolons. Columns have precedence over rows. For example, the notation $[A, B; C, D]$ corresponds to the matrix $\begin{pmatrix} A & B \\ C & D \end{pmatrix}$.

As we show later, the presence of the variables $w, z$ in the objective function may prevent a crucial function from being convex. Since this has an unfavorable impact on the properties of the bilinear program, we introduce a compact form of the problem.

**Definition 2.** We say that the bilinear program in Eq. (1) is in a *compact* form when $s_1$ and $s_2$ are zero vectors. It is in a *semi-compact* form if $s_2$ is a zero vector.

The compactness requirement is not limiting because any bilinear program in the form shown in Eq. (1) can be expressed in a semi-compact form as follows:

$$
\begin{aligned}
\underset{w,x,y,z,\hat{x},\hat{y}}{\text{maximize}} \quad & s_1^\mathsf{T} w + r_1^\mathsf{T} x + \begin{pmatrix} x^\mathsf{T} & \hat{x} \end{pmatrix} \begin{pmatrix} C & 0 \\ 0 & 1 \end{pmatrix} \begin{pmatrix} y \\ \hat{y} \end{pmatrix} + r_2^\mathsf{T} y \\
\text{subject to} \quad & A_1 x + B_1 w = b_1 \quad A_2 y + B_2 z = b_2 \\
& \hat{x} = 1 \quad \hat{y} = s_2^\mathsf{T} z \\
& w, x, y, z \geq \mathbf{0}
\end{aligned}
\tag{4}
$$

Clearly, feasible solutions of Eq. (1) and Eq. (4) have the same objective value when $\hat{y}$ is set appropriately. Notice that the dimensionality of the bilinear term in the objective function increases by 1 for both $x$ and $y$. Hence, this transformation increases the dimensionality of the program by 1.

The rest of this section describes several classes of multiagent planning problems that can be formulated as bilinear programs. Starting with observation and transition independent DEC-MDPs, we extend the formulation to allow a different objective function (maximizing average reward over an infinite horizon), to handle interdependent observations, and to find Nash equilibria in competitive settings.

## 2.1 DEC-MDPs

As mentioned previously, any transition-independent and observation-independent DEC-MDP (Becker et al., 2004) may be formulated as a bilinear program. Intuitively, a DEC-MDP is transition independent when no agent can influence the other agents' transitions. A DEC-MDP is observation independent when no agent can observe the states of other agents. These assumptions are crucial since they ensure a lower complexity of the problem (Becker et al., 2004). In the remainder of the paper, we use simply the term DEC-MDP to refer to transition and observation independent DEC-MDP.

The DEC-MDP model has proved useful in several multiagent planning domains. One example that we use is the Mars rover planning problem (Bresina, Golden, Smith, & Washington, 1999), first formulated as a DEC-MDP by Becker et al. (2003). This domain involves two autonomous rovers that visit several sites in a given order and may decide to perform certain scientific experiments at each site. The overall activity must be completed within a given time limit. The uncertainty about the duration of each experiment is modeled by a given discrete distribution. While the rovers operate independently and receive local rewards for each completed experiment, the global reward function also depends on some experiments completed by both rovers. The interaction between the rovers is thus limited to a relatively small number of such overlapping tasks. We return to this problem and describe it in more detail in Section 6.

A DEC-MDP problem is composed of two MDPs with state-sets $\mathcal{S}_1$, $\mathcal{S}_2$ and action sets $\mathcal{A}_1$, $\mathcal{A}_2$. The functions $r_1$ and $r_2$ define *local* rewards for action-state pairs. The initial state distributions are $\alpha_1$ and $\alpha_2$. The MDPs are coupled through a global reward function defined by the matrix $R$. Each entry $R(i,j)$ represents the joint reward for the state-action $i$ by one agent and $j$ by the other. Our definition of a DEC-MDP is based on the work of Becker et al. (2004), with some modifications that we discuss below.

**Definition 3.** A two-agent transition and observation independent *DEC-MDP* with *extended reward structure* is defined by a tuple $\langle \mathcal{S}, \mathcal{F}, \alpha, \mathcal{A}, P, \mathcal{R} \rangle$:

- $\mathcal{S} = (\mathcal{S}_1, \mathcal{S}_2)$ is the factored set of world states
- $\mathcal{F} = (\mathcal{F}_1 \subseteq \mathcal{S}_1, \mathcal{F}_2 \subseteq \mathcal{S}_2)$ is the factored set of terminal states.
- $\alpha = (\alpha_1, \alpha_2)$ where $\alpha_i : \mathcal{S}_i \mapsto [0,1]$ are the initial state distribution functions
- $\mathcal{A} = (\mathcal{A}_1, \mathcal{A}_2)$ is the factored set of actions
- $P = (P_1, P_2), P_i : \mathcal{S}_i \times \mathcal{A}_i \times \mathcal{S}_i \mapsto [0,1]$ are the transition functions. Let $a \in \mathcal{A}_i$ be an action, then $P_i^a : \mathcal{S}_i \times \mathcal{S}_i \mapsto [0,1]$ is a stochastic transition matrix such that $P_i(s, a, s') = P_i^a(s, s')$ is the probability of a transition from state $s \in \mathcal{S}_i$ to state $s' \in \mathcal{S}_i$ of agent $i$, assuming it takes action $a$. The transitions from the final states have 0 probability; that is $P_i(s, a, s') = 0$ if $s \in \mathcal{F}_i$, $s' \in \mathcal{S}_i$, and $a \in \mathcal{A}_i$.
- $\mathcal{R} = (r_1, r_2, R)$ where $r_i : \mathcal{S}_i \times \mathcal{A}_i \mapsto \mathbb{R}$ are the local reward functions and $R : (\mathcal{S}_1 \times \mathcal{A}_1) \times (\mathcal{S}_2 \times \mathcal{A}_2) \mapsto \mathbb{R}$ is the global reward function. Local rewards $r_i$ are represented as vectors, and $R$ is a matrix with $(s^1, a^1)$ as rows and $(s^2, a^2)$ as columns.

Definition 3 differs from the original definition of transition and observation independent DEC-MDP (Becker et al. 2004, Definition 1) in two ways. The modifications allow us to explicitly capture assumptions that are implicit in previous work. First, the individual MDPs in our model are formulated as stochastic shortest-path problems (Bertsekas & Tsitsiklis, 1996). That is, there is no explicit time horizon, but instead some states are terminal. The process stops upon reaching a terminal state. The objective is to maximize the cumulative reward received before reaching the terminal states.

The second modification of the original definition is that Definition 3 generalizes the reward structure of the DEC-MDP formulation, using the extended reward structure. The joint rewards in the original DEC-MDP are defined only for the joint states ($s^1 \in \mathcal{S}_1, s^2 \in \mathcal{S}_2$) visited by both agents simultaneously. That is, if agent 1 visits states $s_1^1, s_2^1$ and agent 2 visits states $s_1^2, s_2^2$, then the reward can only be defined for joint states $(s_1^1, s_1^2)$ and $(s_2^1, s_2^2)$. However, our DEC-MDP formulation with *extended reward structure* also allows the reward to depend on $(s_1^1, s_2^2)$ and $(s_2^1, s_1^2)$, even when they are not visited simultaneously. As a result, the global reward may depend on the history, not only on the current state. Note that this reward structure is more general than what is commonly used in DEC-POMDPs.

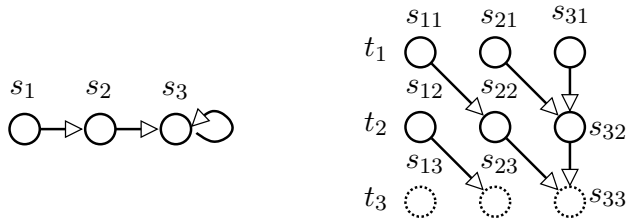

Figure 1: An MDP and its stochastic shortest path version with time horizon 3. The dotted circles are terminal states.

We prefer the more general definition because it has been already implicitly used in previous work. In particular, this extended reward structure arises from introducing the primitive and compound events in the work of Becker et al. (2004). This reward structure is necessary to capture the characteristics of the Mars rover benchmark. Interestingly, this extension does not complicate our proposed solution methods in any way. Note that the stochastic shortest path formulation (right side of Figure 1) inherently eliminates any loops because time always advances when an action is taken. Therefore, every state in that representation may be visited at most once. This property is commonly used when an MDP is formulated as a linear program (Puterman, 2005).

The solution of a DEC-MDP is a deterministic stationary policy $\pi = (\pi_1, \pi_2)$, where $\pi_i : \mathcal{S}_i \mapsto \mathcal{A}_i$ is the standard MDP policy (Puterman, 2005) for agent $i$. In particular, $\pi_i(s^i)$ represents the action taken by agent $i$ in state $s^i$. To define the bilinear program, we use variables $x(s^1, a^1)$ to denote the probability that agent 1 visits state $s^1$ and takes action $a^1$ and $y(s^2, a^2)$ to denote the same for agent 2. These are the standard dual variables in MDP formulation. Given a solution in terms of $x$ for agent 1, the policy is calculated for $s \in \mathcal{S}_1$ as follows, breaking ties arbitrarily.

$$\pi_1(s) = \arg\max_{a \in \mathcal{A}_1} x(s, a)$$

The policy $\pi_2$ is similarly calculated from $y$. The correctness of the policy calculation follows from the existence of an optimal policy that is deterministic and depends only on the local states of that agent (Becker et al., 2004).

The objective in DEC-MDPs in terms of $x$ and $y$ is then to *maximize*:

$$\sum_{\substack{s^1 \in \mathcal{S}_1 \\ a^1 \in \mathcal{A}_1}} r_1(s^1, a^1) x(s^1, a^1) + \sum_{\substack{s^1 \in \mathcal{S}_1 \\ a^1 \in \mathcal{A}_1}} \sum_{\substack{s^2 \in \mathcal{S}_2 \\ a^2 \in \mathcal{A}_2}} R(s^1, a^1, s^2, a^2) x(s^1, a^1) y(s^2, a^2) + \sum_{\substack{s^2 \in \mathcal{S}_2 \\ a^2 \in \mathcal{A}_2}} r_2(s^2, a^2) y(s^2, a^2).$$

The stochastic shortest path representation is more general because any finite-horizon MDP can be represented as such by keeping track of time as part of the state, as illustrated in Figure 1. This modification allows us to apply the model directly to the Mars rover benchmark problem. Actions in the Mars rover problem may have different durations, while all actions in finite-horizon MDPs take the same amount of time.

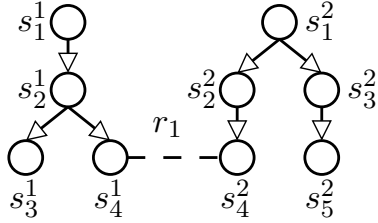

Figure 2: A sample DEC-MDP.

A DEC-MDP problem with an extended reward structure can be formulated as a bilinear mathematical program as follows. Vector variables $x$ and $y$ represent the state-action probabilities for each agent, as used in the dual linear formulation of MDPs. Given the transition and observation independence, the feasible regions may be defined by linear equalities $A_1 x = \alpha_1$ and $x \geq \mathbf{0}$, and $A_2 y = \alpha_2$ and $y \geq \mathbf{0}$. The matrices $A_1$ and $A_2$ are the same as for the dual formulation of total expected reward MDPs (Puterman, 2005), representing the following equalities for agent $i$:

$$\sum_{a' \in \mathcal{A}_i} x(s', a') - \sum_{s \in \mathcal{S}_i} \sum_{a \in \mathcal{A}_i} P_i(s, a, s') x(s, a) = \alpha_i(s'),$$

for every $s' \in \mathcal{S}_i$. As described above, variables $x(s, a)$ represent the probabilities of visiting the state $s$ and taking action $a$ by the appropriate agent during plan execution. Note that for agent 2, the variables are $y(s, a)$ rather than $x(s, a)$. Intuitively, these equalities ensure that the probability of entering each *non-terminal* state, through either the initial step or from other states, is the same as the probability of leaving the state. The bilinear problem is then formulated as follows:

$$\begin{aligned} \underset{x,y}{\text{maximize}} \quad & r_1^\mathsf{T} x + x^\mathsf{T} R y + r_2^\mathsf{T} y \\ \text{subject to} \quad & A_1 x = \alpha_1 \quad x \geq \mathbf{0} \\ & A_2 y = \alpha_2 \quad y \geq \mathbf{0} \end{aligned} \quad (5)$$

In this formulation, we treat the initial state distributions $\alpha_i$ as vectors, based on a fixed ordering of the states. The following simple example illustrates the formulation.

**Example 4.** *Consider a DEC-MDP with two agents, depicted in Figure 2. The transitions in this problem are deterministic, and thus all the branches represent actions $a_i$, ordered for each state from left to right. In some states, only one action is available. The shared reward $r_1$, denoted by a dotted line, is received when both agents visit the state. The local rewards are denoted by the numbers above next to the states. The terminal states are omitted. The* *agents start in states $s_1^1$ and $s_1^2$ respectively. The bilinear formulation of this problem is:*

$$
\begin{aligned}
\text{maximize} \quad & x(s_4^1, a_1) * r_1 * y(s_4^2, a_1) \\
\text{subject to} \quad & x(s_1^1, a_1) = 1 \qquad\qquad y(s_1^2, a_1) + y(s_1^2, a_2) = 1 \\
& x(s_2^1, a_1) + x(s_2^1, a_2) - x(s_1^1, a_1) = 0 \qquad y(s_2^2, a_1) - y(s_1^2, a_1) = 0 \\
& x(s_3^1, a_1) - x(s_2^1, a_1) = 0 \qquad y(s_3^2, a_1) - y(s_2^2, a_1) = 0 \\
& x(s_4^1, a_1) - x(s_2^1, a_2) = 0 \qquad y(s_4^2, a_1) - y(s_2^2, a_1) = 0 \\
& \qquad\qquad\qquad\qquad\qquad\qquad y(s_5^2, a_1) - y(s_3^2, a_1) = 0
\end{aligned}
$$

While the results in this paper focus on two-agent problems, our approach can be extended to DEC-MDPs with more than two agents in two ways. The first approach requires that each component of the global reward depends on at most two agents. The DEC-MDP then may be viewed as a graph with vertices representing agents and edges representing the immediate interactions or dependencies. To formulate the problem as a bilinear program, this graph must be *bipartite*. Interestingly, this class of problems has been previously formulated (Kim et al., 2006). Let $G_1$ and $G_2$ be the indices of the agents in the two partitions of the bipartite graph. Then the problem can be formulated as follows:

$$
\begin{aligned}
\underset{x,y}{\text{maximize}} \quad & \sum_{i \in G_1, j \in G_2} r_i^\mathsf{T} x_i + x_i^\mathsf{T} R_{ij} y_j + r_j^\mathsf{T} y_j \\
\text{subject to} \quad & A_i x_i = \alpha_1 \quad x_i \geq 0 \quad i \in G_1 \\
& A_j y_j = \alpha_2 \quad y_j \geq 0 \quad j \in G_2
\end{aligned}
\tag{6}
$$

Here, $R_{ij}$ denotes the global reward for interactions between agents $i$ and $j$. This program is bilinear and separable because the constraints on the variables in $G_1$ and $G_2$ are independent.

The second approach to generalize the framework is to represent the DEC-MDP as a *multilinear program*. In that case, no restrictions on the reward structure are necessary. An algorithm to solve, say a trilinear program, could be almost identical to the algorithm we propose, except that the best response would be calculated using bilinear, not linear programs. However, the scalability of this approach to more than a few agents is doubtful.

## 2.2 Average-Reward Infinite-Horizon DEC-MDPs

The previous formulation deals with finite-horizon DEC-MDPs. An average-reward problem may also be formulated as a bilinear program (Petrik & Zilberstein, 2007b). This is particularly useful for infinite-horizon DEC-MDPs. For example, consider the infinite-horizon version of the Multiple Access Broadcast Channel (MABC) (Rosberg, 1983; Ooi & Wornell, 1996). In this problem, which has been used widely in recent studies of decentralized decision making, two communication devices share a single channel, and they need to periodically transmit some data. However, the channel can transmit only a single message at a time. When both agents send messages at the same time, this leads to a collision, and the transmission fails. The memory of the devices is limited, thus they need to send the messages sooner rather than later. We adapt the model from the work of Rosberg (1983), which is particularly suitable because it assumes no sharing of local information among the devices.

The definition of average-reward two-agent transition and observation independent DEC-MDP is the same as Definition 3, with the exception of the terminal states, policy, and objective. There are no terminal states in average-reward DEC-MDPs, and the policy $(\pi_1, \pi_2)$ may be stochastic. That is, $\pi_i(s, a) \mapsto [0, 1]$ is the probability of agent $i$ taking an action $a$ in state $s$. The objective is to find a stationary infinite-horizon policy $\pi$ that maximizes the average reward, or *gain*, defined as follows.

**Definition 5.** Let $\pi = (\pi_1, \pi_2)$ be a stochastic policy, and $X_t$ and $Y_t$ be random variables that represent the probability distributions over the state-action pairs at time $t$ of the two agents respectively according to $\pi$. The *gain* $G$ of the policy $\pi$ is then defined for states $s^1 \in \mathcal{S}_1$ and $s^2 \in \mathcal{S}_2$ as:

$$G(s^1, s^2) = \lim_{N \to \infty} \frac{1}{N} \mathbf{E}_{(s^1, \pi_1(s^1)), (s^2, \pi_2(s^2))} \left[ \sum_{t=0}^{N-1} r_1(X_t) + R(X_t, Y_t) + r_2(Y_t) \right],$$

where $\pi_i(s^i)$ is the distribution over the actions in state $s^i$. Note that the expectation is with respect to the initial states and action distributions $(s^1, \pi_1(s^1)), (s^2, \pi_2(s^2))$.

The actual gain of a policy depends on the agents' initial state distributions $\alpha_1, \alpha_2$ and may be expressed as $\alpha_1^\mathsf{T} G \alpha_2$, with $G$ represented as a matrix. Puterman (2005), for example, provides a more detailed discussion of the definition and meaning of policy gain.

To simplify the bilinear formulation of the average-reward DEC-MDP, we assume that $r_1 = \mathbf{0}$ and $r_2 = \mathbf{0}$. The bilinear program follows.

$$
\begin{aligned}
&\underset{p_1, p_2, q_1, q_2}{\text{maximize}} \quad \tau(p_1, p_2, q_1, q_2) = p_1^\mathsf{T} R p_2 \\
&\text{subject to} \quad p_1, p_2 \geq \mathbf{0} \\
&\forall s' \in \mathcal{S}_1 \quad \sum_{a \in \mathcal{A}_1} p_1(s', a) - \sum_{s \in \mathcal{S}_1, a \in \mathcal{A}_1} p_1(s, a) P_1^a\left(s, s'\right) = 0 \\
&\forall s' \in \mathcal{S}_1 \quad \sum_{a \in \mathcal{A}_1} p_1(s', a) + \sum_{a \in \mathcal{A}_1} q_1(s', a) - \sum_{s \in \mathcal{S}_1, a \in \mathcal{A}_1} q_1(s, a) P_1^a\left(s, s'\right) = \alpha_1(s') \\
&\forall s' \in \mathcal{S}_2 \quad \sum_{a \in \mathcal{A}_2} p_2(s', a) - \sum_{s \in \mathcal{S}_2, a \in \mathcal{A}_2} p_2(s, a) P_2^a\left(s, s'\right) = 0 \\
&\forall s' \in \mathcal{S}_2 \quad \sum_{a \in \mathcal{A}_2} p_2(s', a) + \sum_{a \in \mathcal{A}_2} q_2(s', a) - \sum_{s \in \mathcal{S}_2, a \in \mathcal{A}_2} q_2(s, a) P_2^a\left(s, s'\right) = \alpha_2(s')
\end{aligned}
\tag{7}
$$

The variables in the program come from the dual formulation of the average-reward MDP linear program (Puterman, 2005). The state sets of the MDPs is divided into recurrent and transient states. The recurrent states are expected to be visited infinitely many times, while the transient states are expected to be visited finitely many times. Variables $p_1$ and $p_2$ represent the limiting distributions of each MDP, which is non-zero for all recurrent states. The (possibly stochastic) policy $\pi_i$ of agent $i$ is defined in the recurrent states by the probability of taking action $a \in \mathcal{A}_i$ in state $s \in \mathcal{S}_i$:

$$\pi_i(s, a) = \frac{p_i(s, a)}{\sum_{a' \in \mathcal{A}_i} p_i(s, a')}.$$

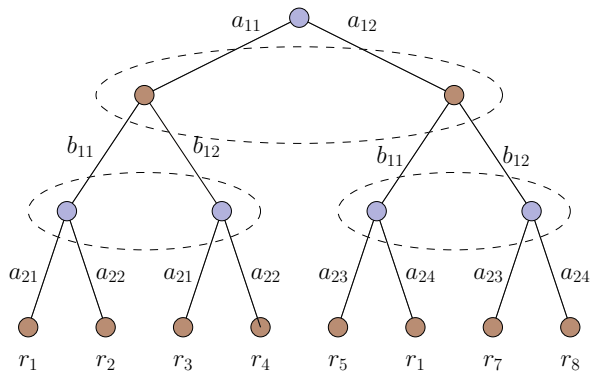

Figure 3: A tree form of a policy for a DEC-POMDP or an extensive game. The dotted ellipses denote the information sets.

The variables $p_i$ are 0 in transient states. The policy in the transient states is calculated from variables $q_i$ as:

$$\pi_i(s, a) = \frac{q_i(s, a)}{\sum_{a' \in \mathcal{A}_i} q_i(s, a')}.$$

The correctness of the constraints follows from the dual formulation of optimal average reward (Puterman 2005, Equation 9.3.4). Petrik and Zilberstein (2007b) provide further details of this formulation.

### 2.3 General DEC-POMDPs and Extensive Games

The general DEC-POMDP problem and extensive-form games with two agents, or players, can also be formulated as bilinear programs. However, the constraints may not be separable because actions of one agent influence the other agent. The approach in this case may be similar to linear complementarity problem formulation of extensive games (Koller, Megiddo, & von Stengel, 1994), and integer linear program formulation of DEC-POMDPs (Aras & Charpillet, 2007). The approach we develop is closely related to event-driven DEC-POMDPs (Becker et al., 2004), but it is in general more efficient. Nevertheless, the size of the bilinear program is exponential in the size of the DEC-POMDP. This can be expected since solving DEC-POMDPs is NEXP-complete (Bernstein et al., 2000), while solving bilinear programs is NP-complete (Mangasarian, 1995). Because the general formulation in this case is somewhat cumbersome, we only illustrate it using the following simple example. Aras (2008) provides the details of a similar construction.

**Example 6.** *Consider the problem depicted in Figure 3, assuming that the agents are cooperative. The actions of the other agent are not observable, as denoted by the information sets. This approach can be generalized to any problem with any observable sets as long as the* perfect recall *condition is satisfied. Agents satisfy the perfect recall condition when they remember the set of actions taken in the prior moves (Osborne & Rubinstein, 1994). Rewards are only collected in the leaf-nodes in this case. The variables on the edges represent the probability of taking the action. Here, variables a denote the actions of one agent, and* *variables $b$ of the other. The total common reward received in the end is:*

$$r = a_{11}b_{11}a_{21}r_1 + a_{11}b_{11}a_{22}r_2 + a_{11}b_{12}a_{21}r_3 + a_{11}b_{12}a_{22}r_4 +$$
$$a_{12}b_{11}a_{23}r_5 + a_{12}b_{11}a_{24}r_6 + a_{12}b_{12}a_{23}r_7 + a_{12}b_{12}a_{24}r_8.$$

*The constraints in this problem are of the following form:* $a_{11} + a_{12} = 1$.

Any DEC-POMDP problem can be represented using the approach used above. It is also straightforward to extend the approach to problems with rewards in every node. However, the above formulation is clearly not bilinear. To apply our algorithm to this class of problems, we need to reformulate the problem in a bilinear form. This can be easily accomplished in a way similar to the construction of the dual linear program for an MDP. Namely, we introduce variables:

$$c_{11} = a_{11}$$
$$c_{12} = a_{12}$$
$$c_{21} = a_{11}a_{21}$$
$$c_{22} = a_{11}a_{22}$$

and so on for every set of variables on any path to a leaf node. Then, the objective may be reformulated as follows:

$$r = c_{21}b_{11}r_1 + c_{22}b_{11}r_2 + c_{23}b_{12}r_3 + c_{24}b_{12}r_4 +$$
$$c_{25}b_{11}r_5 + c_{26}b_{11}r_6 + c_{27}b_{12}r_7 + c_{28}b_{12}r_8.$$

Variables $b_{ij}$ are replaced in the same fashion. This objective function is clearly bilinear. The constraints may be reformulated as follows. The constraint $a_{21} + a_{22} = 1$ can be multiplied by $a_{11}$ and then replaced by $c_{21} + c_{22} = c_{11}$, and so on. That is, the variables in each level have to sum to the variable that is their least common parent in the level above for the same agent.

## 2.4 General Two-Player Games

In addition to cooperative problems, some competitive problems with 2 players may be formulated as bilinear programs. It is known that the problem of finding an equilibrium for a bi-matrix game may be formulated as a linear complementarity problem (Cottle, Pang, & Stone, 1992). It has also been shown that a linear complementarity problem may be formulated as a bilinear problem (Mangasarian, 1995). However, a direct application of these two reductions results in a complex problem with a large dimensionality. Below, we demonstrate how a general game can be directly formulated as a bilinear program. There are many ways to formulate a game, thus we take a very general approach. We simply assume that each agent optimizes a linear program, as follows.

$$\begin{array}{ll} \underset{x}{\text{maximize}} & d_1(x) = r_1^{\mathsf{T}}x + x^{\mathsf{T}}C_1 y \\ \text{subject to} & A_1 x = b_1 \\ & x \geq \mathbf{0} \end{array} \qquad (8)$$

$$\begin{array}{ll} \underset{y}{\text{maximize}} & d_2(y) = r_2^{\mathsf{T}}y + x^{\mathsf{T}}C_2 y \\ \text{subject to} & A_2 y = b_2 \\ & y \geq \mathbf{0} \end{array} \qquad (9)$$

In Eq. (8), the variable $y$ is considered to be a constant and similarly in Eq. (9) the variable $x$ is considered to be a constant. For normal form games, the constraint matrices $A_1$ and $A_2$ are simply rows of ones, and $b_1 = b_2 = 1$. For competitive DEC-MDPs, the constraint matrices $A_1$ and $A_2$ are the same as in Section 2.1. Extensive games may be formulated similarly to DEC-POMDPs, as described in Section 2.3.

The game specified by linear programs Eq. (8) and Eq. (9) may be formulated as a bilinear program as follows. First, define the reward vectors for each agent, given a policy of the other agent.

$$
\begin{aligned}
q_1(y) &= r_1 + C_1 y \\
q_2(x) &= r_2 + C_2^\mathsf{T} x.
\end{aligned}
$$

These values are unrelated to those of Eq. (7). The complementary slackness values (Vanderbei, 2001) for the linear programs Eq. (8) and Eq. (9) are:

$$
\begin{aligned}
k_1(x, y, \lambda_1) &= \left( q_1(y)^\mathsf{T} - \lambda_1^\mathsf{T} A_1 \right) x \\
k_2(x, y, \lambda_2) &= \left( q_2(x)^\mathsf{T} - \lambda_2^\mathsf{T} A_2 \right) y,
\end{aligned}
$$

where $\lambda_1$ and $\lambda_2$ are the dual variables of the corresponding linear programs. For any primal feasible $x$ and $y$, and dual feasible $\lambda_1$ and $\lambda_2$, we have that $k_1(x, y, \lambda_1) \geq 0$ and $k_2(x, y, \lambda_2) \geq 0$. The equality is attained if and only if $x$ and $y$ are optimal. This can be used to write the following optimization problem, in which we implicitly assume that $x, y, \lambda_1, \lambda_2$ are feasible in the appropriate primal and dual linear programs:

$$
\begin{aligned}
0 \leq{}& \min_{x, y, \lambda_1, \lambda_2} k_1(x, y, \lambda_1) + k_2(x, y, \lambda_2) \\
={}& \min_{x, y, \lambda_1, \lambda_2} (q_1(y)^\mathsf{T} - \lambda_1^\mathsf{T} A_1) x + (q_2(x)^\mathsf{T} - \lambda_2^\mathsf{T} A_2) y \\
={}& \min_{x, y, \lambda_1, \lambda_2} ((r_1 + C_1 y)^\mathsf{T} - \lambda_1^\mathsf{T} A_1) x + ((r_2 + C_2^\mathsf{T} x)^\mathsf{T} - \lambda_2^\mathsf{T} A_2) y \\
={}& \min_{x, y, \lambda_1, \lambda_2} r_1^\mathsf{T} x + r_2^\mathsf{T} y + x^\mathsf{T}(C_1 + C_2) y - x^\mathsf{T} A_1^\mathsf{T} \lambda_1 - y^\mathsf{T} A_2^\mathsf{T} \lambda_2 \\
={}& \min_{x, y, \lambda_1, \lambda_2} r_1^\mathsf{T} x + r_2^\mathsf{T} y + x^\mathsf{T}(C_1 + C_2) y - b_1^\mathsf{T} \lambda_1 - b_2^\mathsf{T} \lambda_2.
\end{aligned}
$$

Therefore, any feasible $x$ and $y$ that set the right hand side to 0 solve both linear programs in Eq. (8) and Eq. (9) optimally. Adding the primal and dual feasibility conditions to the above, we get the following bilinear program:

$$
\begin{aligned}
\underset{x, y, \lambda_1, \lambda_2}{\text{minimize}} \quad & r_1^\mathsf{T} x + r_2^\mathsf{T} y + x^\mathsf{T}(C_1 + C_2) y - b_1^\mathsf{T} \lambda_1 - b_2^\mathsf{T} \lambda_2 \\
\text{subject to} \quad & A_1 x = b_1 \quad A_2 y = b_2 \\
& r_1 + C_1 y - A_1^\mathsf{T} \lambda_1 \leq \mathbf{0} \\
& r_2 + C_2^\mathsf{T} x - A_2^\mathsf{T} \lambda_2 \leq \mathbf{0} \\
& x \geq \mathbf{0} \quad y \geq \mathbf{0}
\end{aligned}
\tag{10}
$$

---

**Algorithm 1**: IterativeBestResponse($\mathcal{B}$)

**1** $x_0, w_0 \leftarrow$ rand ;
**2** $i \leftarrow 1$ ;
**3 while** $y_{i-1} \neq y_i$ *or* $x_{i-1} \neq x_i$ **do**
**4**     $(y_i, z_i) \leftarrow \arg\max_{y,z} f(w_{i-1}, x_{i-1}, y, z)$ ;
**5**     $(x_i, w_i) \leftarrow \arg\max_{x,w} f(w, x, y_i, z_i)$ ;
**6**     $i \leftarrow i + 1$
**7 return** $f(w_i, x_i, y_i, z_i)$

---

The optimal solution of Eq. (10) is 0 and it corresponds to a Nash equilibrium. This is because both the primal variables $x, y$ and dual variables $\lambda_1, \lambda_2$ are feasible and the complementary slackness condition is satisfied. The open question in this example are the interpretation of an approximate result and a formulation that would select the equilibrium. It is not clear yet whether it is possible to formulate the program so that the optimal solution will be a Nash equilibrium that maximizes a certain criterion. The approximate solutions of the program probably correspond to $\epsilon$-Nash equilibria, but this remain an open question.

The algorithm in this case also relies on the number of shared rewards being small compared to the size of the problem. But even if this is not the case, it is often possible that the number of shared rewards may be automatically reduced as described in Section 4. In fact, it is easy to show that a zero-sum normal form game is automatically reduced to two uncoupled linear programs. This follows from the dimensionality reduction procedure in Section 4.

## 3. Solving Bilinear Programs

One simple method often used for solving bilinear programs is the iterative procedure shown in Algorithm 1. The parameter $\mathcal{B}$ represents the bilinear program. While the algorithm often performs well in practice, it tends to converge to a suboptimal solution (Mangasarian, 1995). When applied to DEC-MDPs, this algorithm is essentially identical to JESP (Nair, Tambe, Yokoo, Pynadath, & Marsella, 2003)–one of the early solution methods. In the following, we use $f(w, x, y, z)$ to denote the objective value of Eq. (1).

The rest of this section presents a new anytime algorithm for solving bilinear programs. The goal of the algorithm to is to produce a good solution quickly and then improve the solution in the remaining time. Along with each approximate solution, the maximal approximation bound with respect to the optimal solution is provided. As we show below, our algorithm can benefit from results produced by suboptimal algorithms, such as Algorithm 1, to quickly determine tight approximation bounds.

### 3.1 The Successive Approximation Algorithm

We begin with an overview of a successive approximation algorithm for bilinear problems that takes advantage of a low number of interactions between the agents. It is particularly suitable when the input problem is large in comparison to its dimensionality, as defined in Section 2. We address the issue of dimensionality reduction in Section 4.

We begin with a simple intuitive explanation of the algorithm, and then show how it can be formalized. The bilinear program can be seen as an optimization game played by two agents, in which the first agent sets the variables $w, x$ and the second one sets the variables $y, z$. This is a general observation that applies to any bilinear program. In any practical application, the feasible sets for the two sets of variables may be too large to explore exhaustively. In fact, when this method is applied to DEC-MDPs, these sets are infinite and continuous. The basic idea of the algorithm is to first identify the set of best responses of one of the agents, say agent 1, to some policy of the other agent. This is simple because once the variables of agent 2 are fixed, the program becomes linear, which is relatively easy to solve. Once the set of best-response policies of agent 1 is identified, assuming it is of a reasonable size, it is possible to calculate the best response of agent 2.

This general approach is also used by the coverage set algorithm (Becker et al., 2004). One distinction is that the representation used in CSA applies only to DEC-MDPs, while our formulation applies to bilinear programs–a more general representation. The main distinction between our algorithm and CSA is the way in which the variables $y, z$ are chosen. In CSA, the values $y, z$ are calculated in a way that simply guarantees termination in finite time. We, on the other hand, choose values $y, z$ greedily so as to minimize the approximation bound on the optimal solution. This is possible because we establish bounds on the optimality of the solution throughout the calculation. As a result, our algorithm converges more rapidly and may be terminated at any time with a guaranteed performance bound. Unlike the earlier version of the algorithm (Petrik & Zilberstein, 2007a), the version described in this paper calculates the best response using only a subset of the values of $y, z$. As we show, it is possible to identify regions of $y, z$ in which it is impossible to improve the current best solution and exclude these regions from consideration.

We now formalize the ideas described above. To simplify the notation, we define feasible sets as follows:

$$
\begin{aligned}
X &= \{(x, w) \,|\, A_1 x + B_1 w = b_1\} \\
Y &= \{(y, z) \,|\, A_2 y + B_2 z = b_2\}.
\end{aligned}
$$

We use $y \in Y$ to denote that there exists $z$ such that $(y, z) \in Y$. In addition, we assume that the problem is in a semi-compact form. This is reasonable because any bilinear program may be converted to semi-compact form with an increase in dimensionality of one, as we have shown earlier.

**Assumption 7.** The sets $X$ and $Y$ are bounded, that is, they are contained in a ball of a finite radius.

While Assumption 7 is limiting, coordination problems under uncertainty typically have bounded feasible sets because the variables correspond to probabilities bounded to $[0, 1]$.

**Assumption 8.** The bilinear program is in a semi-compact form.

The main idea of the algorithm is to compute a set $\tilde{X} \subseteq X$ that contains only those elements that satisfy a necessary optimality condition. The set $\tilde{X}$ is formally defined as follows:

$$
\tilde{X} \subseteq \left\{ (x^*, w^*) \,\middle|\, \exists (y, z) \in Y \; f(w^*, x^*, y, z) = \max_{(x, w) \in X} f(w, x, y, z) \right\}.
$$

As described above, this set may be seen as a set of best responses of one agent to the variable settings of the other. The best responses are easy to calculate since the bilinear program in Eq. (1) reduces to a linear program for fixed $w, x$ or fixed $y, z$. In our algorithm, we assume that $\tilde{X}$ is potentially a proper subset of all necessary optimality points and focus on the approximation error of the optimal solution. Given the set $\tilde{X}$, the following simplified problem is solved.

$$
\begin{aligned}
\underset{w,x,y,z}{\text{maximize}} \quad & f(w,x,y,z) \\
\text{subject to} \quad & (x,w) \in \tilde{X} \\
& A_2 y + B_2 z = b_2 \\
& y, z \geq \mathbf{0}
\end{aligned}
\tag{11}
$$

Unlike the original continuous set $X$, the reduced set $\tilde{X}$ is discrete and small. Thus the elements of $\tilde{X}$ may be enumerated. For a fixed $w$ and $x$, the bilinear program in Eq. (11) reduces to a linear program.

To help compute the approximation bound and to guide the selection of elements for $\tilde{X}$, we use the *best-response function* $g(y)$, defined as follows:

$$
g(y) = \max_{\{w,x,z \mid (x,w) \in X, (y,z) \in Y\}} f(w,x,y,z) = \max_{\{x,w \mid (x,w) \in X\}} f(w,x,y,0),
$$

with the second equality for semi-compact programs only and feasible $y \in Y$. Note that $g(y)$ is also defined for $y \notin Y$, in which case the choice of $z$ is arbitrary since it does not influence the objective function. The best-response function is easy to calculate using a linear program. The crucial property of the function $g$ that we use to calculate the approximation bound is its *convexity*. The following proposition holds because $g(y) = \max_{\{x,w \mid (x,w) \in X\}} f(w,x,y,0)$ is a maximum of a finite set of linear functions.

**Proposition 9.** *The function $g(y)$ is convex when the program is in a semi-compact form.*

Proposition 9 relies heavily on the separability of Eq. (1), which means that the constraints on the variables on one side of the bilinear term are independent of the variables on the other side. The separability ensures that $w, x$ are valid solutions regardless of the values of $y, z$. The semi-compactness of the program is necessary to establish convexity, as shown in Example 23 in Appendix C. The example is constructed using the properties described in the appendix, which show that $f(w,x,y,z)$ may be expressed as a sum of a convex and a concave function.

We are now ready to describe Algorithm 2, which computes the set $\tilde{X}$ for a bilinear problem $\mathcal{B}$ such that the approximation error is at most $\epsilon_0$. The algorithm iteratively adds the best response $(x,w)$ for a selected *pivot point* $y$ into $\tilde{X}$. The pivot points are selected hierarchically. At an iteration $j$, the algorithm keeps a set of polyhedra $S_1 \ldots S_j$ which represent the triangulation of the feasible space $Y$, which is possible based on Assumption 7. For each polyhedron $S_i = (y_1 \ldots y_{n+1})$, the algorithm keeps a bound $\epsilon_i$ on the maximal difference between the optimal solution on the polyhedron and the best solution found so far. This error bound on a polyhedron $S_i$ is defined as:

$$
\epsilon_i = e(S_i) = \max_{\{w,x,y \mid (x,w) \in X, y \in S_i\}} f(w,x,y,0) - \max_{\{w,x,y \mid (x,w) \in \tilde{X}, y \in S_i\}} f(w,x,y,0),
$$

---

**Algorithm 2**: BestResponseApprox($\mathcal{B}, \epsilon_0$) returns $(w, x, y, z)$

---

    // Create the initial polyhedron $S_1$.

**1** $S_1 \leftarrow (y_1 \ldots y_{n+1}), Y \subseteq S_1$ ;

    // Add best-responses for vertices of $S_1$ to $\tilde{X}$

**2** $\tilde{X} \leftarrow \{\arg\max_{(x,w) \in X} f(w, x, y_1, 0), ..., \arg\max_{(x,w) \in X} f(w, x, y_{n+1}, 0)\}$ ;

    // Calculate the error $\epsilon$ and pivot point $\phi$ of the initial polyhedron

**3** $(\epsilon_1, \phi_1) \leftarrow PolyhedronError(S_1)$ ;          // Section 3.2,Section 3.3

    // Initialize the number of polyhedra to 1

**4** $j \leftarrow 1$ ;

    // Continue until reaching a predefined precision $\epsilon_0$

**5** **while** $\max_{i=1,\ldots,j} \epsilon_i \geq \epsilon_0$ **do**

        // Find the polyhedron with the largest error

**6**     $i \leftarrow \arg\max_{k=1,\ldots,j} \epsilon_k$ ;

        // Select the pivot point of the polyhedron with the largest error

**7**     $y \leftarrow \phi_i$ ;

        // Add the best response to the pivot point $y$ to the set $\tilde{X}$

**8**     $\tilde{X} \leftarrow X \cup \{\arg\max_{(x,w) \in X} f(w, x, y, 0)\}$ ;

        // Calculate errors and pivot points of the refined polyhedra

**9**     **for** $k = 1, \ldots, n+1$ **do**

**10**         $j \leftarrow j + 1$ ;

            // Replace the $k$-th vertex by the pivot point $y$

**11**         $S_j \leftarrow (y, y_1 \ldots y_{k-1}, y_{k+1}, \ldots y_{n+1})$ ;

**12**         $(\epsilon_j, \phi_j) \leftarrow PolyhedronError(S_j)$ ;         // Section 3.2,Section 3.3

            // Take the smaller of the errors on the original and the refined polyhedron. The error may not increase with the refinement, although the bound may

**13**         $\epsilon_j \leftarrow \min\{\epsilon_i, \epsilon_j\}$ ;

        // Set the error of the refined polyhedron to 0, since the region is covered by the refinements

**14**     $\epsilon_i \leftarrow 0$ ;

**15** $(w, x, y, z) \leftarrow \arg\max_{\{w,x,y,z \mid (x,w) \in \tilde{X}, (y,z) \in Y\}} f(w, x, y, 0)$ ;

**16** **return** $(w, x, y, z)$ ;

---

where $\tilde{X}$ represents the current, not final, set of best responses.

Next, a point $y_0$ is selected as described below and $n+1$ new polyhedra are created by replacing one of the vertices by $y_0$ to get: $(y_0, y_2, \ldots), (y_1, y_0, y_3, \ldots), \ldots, (y_1, \ldots, y_n, y_0)$. This is depicted for a 2-dimensional set $Y$ in Figure 4. The old polyhedron is discarded and the above procedure is then repeatedly applied to the polyhedron with the maximal approximation error.

For the sake of clarity, the pseudo-code of Algorithm 2 is simplified and does not address any efficiency issues. In practice, $g(y_i)$ could be cached, and the errors $\epsilon_i$ could be stored in a prioritized heap or at least in a sorted array. In addition, a lower bound $l_i$ and an upper bound $u_i$ is calculated and stored for each polyhedron $S_i = (y_1 \ldots y_{n+1})$. The function $e(S_i)$ calculates their maximal difference on the polyhedron $S_i$ and the point where it is attained. The error bound $\epsilon_i$ on the polyhedron $S_i$ may not be tight, as we describe in Remark 13. As a result, when the polyhedron $S_i$ is refined to $n$ polyhedra $S'_1 \ldots S'_n$ with online error bounds $\epsilon'_1 \ldots \epsilon'_n$, it is possible that for some $k$: $\epsilon'_k > \epsilon_i$. Since $S'_1 \ldots S'_n \subseteq S_i$, the true error on $S'_k$ is less than on $S_i$ and therefore $\epsilon'_k$ may be set to $\epsilon_i$.

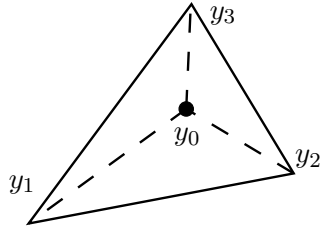

Figure 4: Refinement of a polyhedron in two dimensions with a pivot $y_0$.

Conceptually, the algorithm is similar to CSA, but there are some important differences. The main difference is in the choice of the pivot point $y_0$ and the bounds on $g$. CSA does not keep any upper bound and it evaluates $g(y)$ on all the intersection points of planes defined by the current solutions in $\tilde{X}$. That guarantees that $g(y)$ is eventually known precisely (Becker et al., 2004). A similar approach was also taken for POMDPs (Cheng, 1988). The upper bound on the number of intersection points in CSA is $\binom{|\tilde{X}|}{\dim Y}$. The principal problem is that the bound is exponential in the dimension of $Y$, and experiments do not show a slower growth in typical problems. In contrast, we choose the pivot points to minimize the approximation error. This is more selective and tends to more rapidly reduce the error bound. In addition, the error at the pivot point may be used to determine the overall error bound. The following proposition states the soundness of the triangulation, proved in Appendix A. The correctness of the triangulation establishes that in each iteration the approximation error over $Y$ is equivalent to the maximum of the approximation errors over the current polyhedra $S_1 \ldots S_j$.

**Proposition 10.** *In the proposed triangulation, the sub-polyhedra do not overlap and they cover the whole feasible set $Y$, given that the pivot point is in the interior of $S$.*

### 3.2 Online Error Bound

The selection of the pivot point plays a key role in the performance of the algorithm, in both calculating the error bound and the speed of convergence to the optimal solution. In this section we show exactly how we use the triangulation in the algorithm to calculate an error bound. To compute the approximation bound, we define the *approximate best-response function $\tilde{g}(y)$* as:

$$\tilde{g}(y) = \max_{\{x,w \,|\, (x,w) \in \tilde{X}\}} f(w, x, y, 0).$$

Notice that $z$ is not considered in this expression, since we assume that the bilinear program is in the semi-compact form. The value of the best approximate solution during the execution of the algorithm is:

$$\max_{\{w,x,y,z \,|\, (x,w) \in \tilde{X}, y \in Y\}} f(w, x, y, 0) = \max_{y \in Y} \tilde{g}(y).$$

This value can be calculated at runtime when each new element of $\tilde{X}$ is added. Then the maximal approximation error between the current solution and the optimal one may be calculated from the approximation error of the best-response function $g(\cdot)$, as stated by the following proposition.

**Proposition 11.** *Consider a bilinear program in a semi-compact form. Then let $\tilde{w}, \tilde{x}, \tilde{y}$ be an optimal solution of Eq. (11) and let $w^*, x^*, y^*$ be an optimal solution of Eq. (1). The approximation error is then bounded by:*

$$f(w^*, x^*, y^*, 0) - f(\tilde{w}, \tilde{x}, \tilde{y}, 0) \leq \max_{y \in Y} \left( g(y) - \tilde{g}(y) \right).$$

*Proof.*

$$f(w^*, x^*, y^*, 0) - f(\tilde{w}, \tilde{x}, \tilde{y}, 0) = \max_{y \in Y} g(y) - \max_{y \in Y} \tilde{g}(y) \leq \max_{y \in Y} g(y) - \tilde{g}(y)$$

$\square$

Now, the approximation error is $\max_{y \in Y} g(y) - \tilde{g}(y)$, which is bounded by the difference between an upper bound and a lower bound on $g(y)$. Clearly, $\tilde{g}(y)$ is a lower bound on $g(y)$. Given points in which $\tilde{g}(y)$ is the same as the best-response function $g(y)$, we can use Jensen's inequality to obtain the upper bound. This is summarized by the following lemma.

**Lemma 12.** *Let $y_i \in Y$ for $i = 1, \ldots, n+1$ such that $\tilde{g}(y_i) = g(y_i)$. Then $g\left( \sum_{i=1}^{n+1} c_i y_i \right) \leq \sum_{i=1}^{n+1} c_i g(y_i)$ when $\sum_{i=1}^{n+1} c_i = 1$ and $c_i \geq 0$ for all $i$.*

The actual implementation of the bound relies on the choice of the pivot points. Next we describe the maximal error calculation on a single polyhedron defined by $S = (y_1 \ldots y_n)$. Let matrix $T$ have $y_i$ as columns, and let $L = \{x_1 \ldots x_{n+1}\}$ be the set of the best responses for its vertices. The matrix $T$ is used to convert any $y$ in absolute coordinates to a relative representation $t$ that is a convex combination of the vertices. This is defined formally as follows:

$$y = Tt = \begin{pmatrix} | & | & \ldots \\ y_1 & y_2 & \ldots \\ | & | & \ldots \end{pmatrix} t$$

$$1 = \mathbf{1}^\mathsf{T} t$$

$$0 \leq t$$

where the $y_i$'s are column vectors.

We can represent a lower bound $l(y)$ for $\tilde{g}(y)$ and an upper bound $u(y)$ for $g(y)$ as:

$$l(y) = \max_{x \in L} r^\mathsf{T} x + x^\mathsf{T} C y$$

$$u(y) = [g(y_1), g(y_2), \ldots]^\mathsf{T} t = [g(y_1), g(y_2), \ldots]^\mathsf{T} \begin{pmatrix} T \\ \mathbf{1}^\mathsf{T} \end{pmatrix}^{-1} \begin{pmatrix} y \\ 1 \end{pmatrix},$$

The upper bound correctness follows from Lemma 12. Notice that $u(y)$ is a linear function, which enables us to use a linear program to determine the maximal-error point.

---

**Algorithm 3**: PolyhedronError($\mathcal{B}$, $S$)

---

**1** $\mathcal{P} \leftarrow$ one of Eq. (12), or (13), or (14), or (20) ;

**2** $t \leftarrow$ the optimal solution of $\mathcal{P}$ ;

**3** $\epsilon \leftarrow$ the optimal objective value of $\mathcal{P}$ ;

      // Coordinates $t$ are relative to the vertices of $S$, convert them to absolute values in $Y$

**4** $\phi \leftarrow Tt$ ;

**5** **return** $(\epsilon, \phi)$ ;

---

*Remark* 13. Notice that we use $L$ instead of $\tilde{X}$ in calculating $l(y)$. Using all of $\tilde{X}$ would lead to a tighter bound, as it is easy to show in three-dimensional examples. However, this also would substantially increase the computational complexity.

Now, the error on a polyhedron $S$ may be expressed as:

$$
\begin{aligned}
e(S) &\leq \max_{y \in S} u(y) - l(y) = \max_{y \in S} u(y) - \max_{x \in L} r^{\mathsf{T}}x + x^{\mathsf{T}}Cy \\
&= \max_{y \in S} \min_{x \in L} u(y) - r^{\mathsf{T}}x - x^{\mathsf{T}}Cy.
\end{aligned}
$$

We also have

$$
y \in S \Leftrightarrow \left( y = Tt \wedge t \geq \mathbf{0} \wedge \mathbf{1}^{\mathsf{T}}t = 1 \right).
$$

As a result, the point with the maximal error bound may be determined using the following linear program in terms of variables $t, \epsilon$:

$$
\begin{aligned}
\underset{t,\epsilon}{\text{maximize}} \quad & \epsilon \\
\text{subject to} \quad & \epsilon \leq u(Tt) - r^{\mathsf{T}}x - x^{\mathsf{T}}CTt \quad \forall x \in L \\
& \mathbf{1}^{\mathsf{T}}t = 1 \quad t \geq \mathbf{0}
\end{aligned}
\tag{12}
$$

Here $x$ is *not* a variable. The formulation is correct because all feasible solutions are bounded below the maximal error and any maximal-error solution is feasible.

**Proposition 14.** *The optimal solution of Eq. (12) is equivalent to* $\max_{y \in S} |u(y) - l(y)|$.

We thus select the next pivot point to greedily minimize the error. The maximal difference is actually achieved in points where some of the planes meet, as Becker et al. (2004) have suggested. However, checking these intersections is very similar to running the simplex algorithm. In general, the simplex algorithm is preferable to interior point methods for this program because of its small size (Vanderbei, 2001).

Algorithm 3 shows a general way to calculate the maximal error and the pivot point on the polyhedron $S$. This algorithm may use the basic formulation in Eq. (12), or the more advanced formulations in Eqs. (13), (14), and (20) defined in Section 3.3.

In the following section, we describe a more refined pivot point selection method that can in some cases dramatically improve the performance.

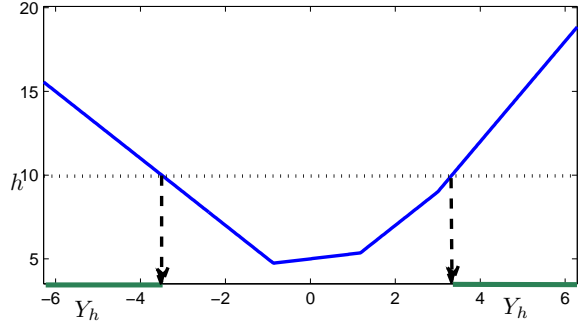

Figure 5: The reduced set $Y_h$ that needs to be considered for pivot point selection.

## 3.3 Advanced Pivot Point Selection

As described above, the pivot points are chosen greedily to both determine the maximal error in each polyhedron and to minimize the approximation error. The basic approach described in Section 3.1 may be refined, because the goal is not to approximate the function $g(y)$ with the least error, but to find the optimal solution. Intuitively, we can ignore those regions of $Y$ that will not guarantee any improvement of the current solution, as illustrated in Figure 5. As we show below, the search for the maximal error point could be limited to this region as well.

We first define a set $Y_h \subseteq Y$ that we will search for the maximal error, given that the optimal solution $f^* \geq h$.

$$Y_h = \{y \mid g(y) \geq h, y \in Y\}.$$

The next proposition states that the maximal error needs to be calculated only in a superset of $Y_h$.

**Proposition 15.** *Let $\tilde{w}, \tilde{x}, \tilde{y}, \tilde{z}$ be the approximate optimal solution and $w^*, x^*, y^*, z^*$ be the optimal solution. Also let $f(w^*, x^*, y^*, z^*) \geq h$ and assume some $\tilde{Y}_h \supseteq Y_h$. The approximation error is then bounded by:*

$$f(w^*, x^*, y^*, z^*) - f(\tilde{w}, \tilde{x}, \tilde{y}, \tilde{z}) \leq \max_{y \in \tilde{Y}_h} g(y) - \tilde{g}(y).$$

*Proof.* First, $f(w^*, x^*, y^*, z^*) = g(y^*) \geq h$ and thus $y^* \in Y_h$. Then:

$$
\begin{aligned}
f(w^*, x^*, y^*, z^*) - f(\tilde{w}, \tilde{x}, \tilde{y}, \tilde{z}) &= \max_{y \in Y_h} g(y) - \max_{y \in Y} \tilde{g}(y) \\
&\leq \max_{y \in Y_h} g(y) - \tilde{g}(y) \\
&\leq \max_{y \in \tilde{Y}_h} g(y) - \tilde{g}(y)
\end{aligned}
$$

$\square$

Proposition 15 indicates that the point with the maximal error needs to be selected only from the set $Y_h$. The question is how to easily identify $Y_h$. Because the set is not convex in general, a tight approximation of this set needs to be found. In particular, we use methods that approximate the intersection of a superset of $Y_h$ with the polyhedron that is being refined, using the following methods:

1. *Feasibility* [Eq. (13)]: Require that pivot points are feasible in $Y$.
2. *Linear bound* [Eq. (14)]: Use the linear upper bound $u(y) \geq h$.
3. *Cutting plane* [Eq. (20)]: Use the linear inequalities that define $Y_h^{\mathsf{C}}$, where $Y_h^{\mathsf{C}} = \mathbb{R}^{|Y|} \setminus Y_h$ is the complement of $Y_h$.

Any combination of these methods is also possible.

**Feasibility**    The first method is the simplest, but also the least constraining. The linear program to find the pivot point with the maximal error bound is as follows:

$$
\begin{aligned}
\underset{\epsilon, t, y, z}{\text{maximize}} \quad & \epsilon \\
\text{subject to} \quad & \epsilon \leq u(Tt) - r^{\mathsf{T}}x + x^{\mathsf{T}}CTt \quad \forall x \in L \\
& \mathbf{1}^{\mathsf{T}}t = 1 \quad t \geq \mathbf{0} \\
& y = Tt \\
& A_2 y + B_2 z = b_2 \\
& y, z \geq \mathbf{0}
\end{aligned}
\tag{13}
$$

This approach does not require that the bilinear program is in the semi-compact form.

**Linear Bound**    The second method, using the linear bound, is also very simple to implement and compute, and it is more selective than just requiring feasibility. Let:

$$
\tilde{Y}_h = \{y \,|\, u(y) \geq h\} \supseteq \{y \,|\, g(y) \geq h\} = Y_h.
$$

This set is convex and thus does not need to be approximated. The linear program used to find the pivot point with the maximal error bound is as follows:

$$
\begin{aligned}
\underset{\epsilon, t}{\text{maximize}} \quad & \epsilon \\
\text{subject to} \quad & \epsilon \leq u(Tt) - r^{\mathsf{T}}x + x^{\mathsf{T}}CTt \quad \forall x \in L \\
& \mathbf{1}^{\mathsf{T}}t = 1 \quad t \geq \mathbf{0} \\
& u(Tt) \geq h
\end{aligned}
\tag{14}
$$

The difference from Eq. (12) is the last constraint. This approach requires that the bilinear program is in the semi-compact form to ensure that $u(y)$ is a bound on the total return.

**Cutting Plane**    The third method, using the cutting plane elimination, is the most computationally intensive one, but also the most selective one. Using this approach requires additional assumptions on the other parts of the algorithm, which we discuss below. The method is based on the same principle as $\alpha$-extensions in concave cuts (Horst & Tuy, 1996). We start with the set $Y_h^{\mathsf{C}}$ because it is convex and may be expressed as:

$$
\left( \max_{w,x} s_1^{\mathsf{T}}w + r_1^{\mathsf{T}}x + y^{\mathsf{T}}C^{\mathsf{T}}x + r_2^{\mathsf{T}}y \right) \quad \leq \quad h \tag{15}
$$

$$
A_1 x + B_1 w \quad = \quad b_1 \tag{16}
$$

$$
w, x \quad \geq \quad \mathbf{0} \tag{17}
$$

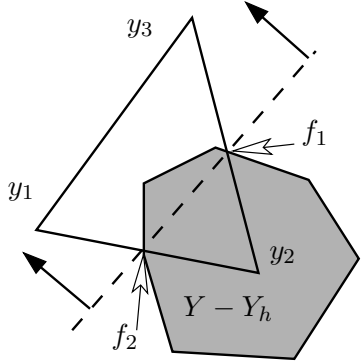

Figure 6: Approximating $Y_h$ using the cutting plane elimination method.

To use these inequalities in selecting the pivot point, we need to make them linear. But there are two obstacles: Eq. (15) contains a bilinear term and is a maximization. Both of these issues can be addressed by using the dual formulation of Eq. (15). The corresponding linear program and its dual for fixed $y$, ignoring constants $h$ and $r_2^\mathsf{T} y$, are:

$$
\begin{array}{ll}
\underset{w,x}{\text{maximize}} & s_1^\mathsf{T} w + r_1^\mathsf{T} x + y^\mathsf{T} C^\mathsf{T} x \\
\text{subject to} & A_1 x + B_1 w = b_1 \\
& w, x \geq \mathbf{0}
\end{array}
\qquad (18)
$$

$$
\begin{array}{ll}
\underset{\lambda}{\text{minimize}} & b_1^\mathsf{T} \lambda \\
\text{subject to} & A_1^\mathsf{T} \lambda \geq r_1 + Cy \\
& B_1^\mathsf{T} \lambda \geq s_1
\end{array}
\qquad (19)
$$

Using the dual formulation, Eq. (15) becomes:

$$
\begin{aligned}
\left( \min_{\lambda} b_1^\mathsf{T} \lambda + r_2^\mathsf{T} y \right) &\leq h \\
A_1^\mathsf{T} \lambda &\geq r_1 + Cy \\
B_1^\mathsf{T} \lambda &\geq s_1
\end{aligned}
$$

Now, we use that for any function $\phi$ and any value $\theta$ the following holds:

$$
\min_x \phi(x) \leq \theta \Leftrightarrow (\exists x) \ \ \phi(x) \leq \theta.
$$

Finally, this leads to the following set of inequalities.

$$
\begin{aligned}
r_2^\mathsf{T} y &\leq h - b_1^\mathsf{T} \lambda \\
Cy &\leq A_1^\mathsf{T} \lambda - r_1 \\
s_1 &\leq B_1^\mathsf{T} \lambda
\end{aligned}
$$

The above inequalities define the convex set $Y_h^{\mathsf{C}}$. Because its complement $Y_h$ is not necessarily convex, we need to use its convex superset $\tilde{Y}_h$ on the given polyhedron. This is done by projecting $Y_h^{\mathsf{C}}$, or its subset, onto the edges of each polyhedron as depicted in Figure 6 and described in Algorithm 4. The algorithm returns a single constraint which cuts off part of the set $Y_h^{\mathsf{C}}$. Notice that only the combination of the first $n$ points $f_k$ is

---

**Algorithm 4**: PolyhedronCut($\{y_1, \ldots, y_{n+1}\}, h$) returns constraint $\sigma^\mathsf{T} y \leq \tau$

---

    `// Find vertices of the polyhedron `$\{y_1, \ldots, y_{n+1}\}$` inside of `$Y_h^\mathsf{C}$

**1**   $\mathcal{I} \leftarrow \{y_i \,|\, y_i \in Y_h^\mathsf{C}\}$ ;

    `// Find vertices of the polyhedron outside of `$Y_h^\mathsf{C}$

**2**   $\mathcal{O} \leftarrow \{y_i \,|\, y_i \in Y_h\}$ ;

    `// Find at least `$n$` points `$f_k$` in which the edge of `$Y_h$` intersects an edge of the polyhedron`

**3**   $k \leftarrow 1$ ;

**4**   **for** $i \in \mathcal{O}$ **do**

**5**      **for** $j \in \mathcal{I}$ **do**

**6**          $f_k \leftarrow y_j + \max_\beta \{\beta \,|\, \beta(y_i - y_j) \in (Y_h^\mathsf{C})\}$ ;

**7**          $k \leftarrow k + 1$ ;

**8**          **if** $k \geq n$ **then**

**9**             **break** ;

**10** Find $\sigma$ and $\tau$, such that $[f_1, \ldots, f_n]\sigma = \tau$ and $\mathbf{1}^\mathsf{T}\sigma = 1$ ;

    `// Determine the correct orientation of the constraint to have all `$y$` in `$Y_h$` feasible`

**11** **if** $\exists y_j \in \mathcal{O}$, and $\sigma^\mathsf{T} y_j > \tau$ **then**

    `// Reverse the constraint if it points the wrong way`

**12**      $\sigma \leftarrow -\sigma$ ;

**13**      $\tau \leftarrow -\tau$ ;

**14** **return** $\sigma^\mathsf{T} y \leq \tau$

---

used. In general, there may be more than $n$ points, and any subset of points $f_k$ of size $n$ can be used to define a new cutting plane that constraints $Y_h$. This did not lead to significant improvements in our experiments. The linear program to find the pivot point with the cutting plane option is as follows:

$$
\begin{aligned}
\underset{\epsilon, t, y}{\text{maximize}} \quad & \epsilon \\
\text{subject to} \quad & \epsilon \leq u(Tt) - r^\mathsf{T}x + x^\mathsf{T}CTt \quad \forall x \in L \\
& \mathbf{1}^\mathsf{T}t = 1 \quad t \geq \mathbf{0} \\
& y = Tt \\
& \sigma^\mathsf{T}y \leq \tau
\end{aligned}
\tag{20}
$$

Here, $\sigma$, and $\tau$ are obtained as a result of running Algorithm 4.

Note that this approach requires that the bilinear program is in the semi-compact form to ensure that $g(y)$ is convex. The following proposition states the correctness of this procedure.

**Proposition 16.** *The resulting polyhedron produced by Algorithm 4 is a superset of the intersection of the polyhedron $S$ with the complement of $Y_h$.*

*Proof.* The convexity of $g(y)$ implies that $Y_h^\mathsf{C}$ is also convex. Therefore, the intersection

$$
Q = \{y \,|\, \sigma^\mathsf{T}y \geq \tau\} \cap S
$$

257

is also convex. It is also a convex hull of points $f_k \in Y_h^{\mathsf{C}}$. Therefore, from the convexity of $Y_h^{\mathsf{C}}$, we have that $Q \subseteq Y_h^{\mathsf{C}}$, and therefore $S - Q \supseteq Y_h$. □

## 4. Dimensionality Reduction

Our experiments show that the efficiency of the algorithm depends heavily on the dimensionality of the matrix $C$ in Eq. (1). In this section, we show the principles behind automatically determining the necessary dimensionality of a given problem. Using the proposed procedure, it is possible to identify weak interactions and eliminate them. Finally, the procedure works for arbitrary bilinear programs and is a generalization of a method we have previously introduced (Petrik & Zilberstein, 2007a).

The dimensionality is inherently part of the model, not the problem itself. There may be equivalent models of a given problem with very different dimensionality. Thus, procedures for reducing the dimensionality are not necessary when the modeler can create a model with minimal dimensionality. However, this is nontrivial in many cases. In addition, some dimensions may have little impact on the overall performance. To determine which ones can be discarded, we need a measure of their contribution that can be computed efficiently. We define these notions more formally later in this section.

We assume that the feasible sets have bounded $L_2$ norms, and assume a general formulation of the bilinear program, not necessarily in the semi-compact form. Given Assumption 7, this can be achieved by scaling the constraints when the feasible region is bounded.

**Assumption 17.** For all $x \in X$ and $y \in Y$, their norms satisfy $\|x\|_2 \leq 1$ and $\|y\|_2 \leq 1$.

We discuss the implications of and problems with this assumption after presenting Theorem 18. Intuitively, the dimensionality reduction removes those dimensions where $g(y)$ is constant, or almost constant. Interestingly, these dimensions may be recovered based on the eigenvectors and eigenvalues of $C^{\mathsf{T}}C$. We use the eigenvectors of $C^{\mathsf{T}}C$ instead of the eigenvectors of $C$, because our analysis is based on $L_2$ norm of $x$ and $y$ and thus of $C$. The $L_2$ norm $\|C\|_2$ is bounded by the largest eigenvalue of $C^{\mathsf{T}}C$. In addition, a symmetric matrix is required to ensure that the eigenvectors are perpendicular and span the whole space.

Given a problem represented using Eq. (1), let $F$ be a matrix whose columns are all the eigenvectors of $C^{\mathsf{T}}C$ with eigenvalues greater than some $\bar{\lambda}$. Let $G$ be a matrix with all the remaining eigenvectors as columns. Notice that together, the columns of the matrices span the whole space and are real-valued, since $C^{\mathsf{T}}C$ is a symmetric matrix. Assume without loss of generality that the eigenvectors are unitary. The *compressed version* of the bilinear program is then the following:

$$
\begin{aligned}
\operatorname*{maximize}_{w,x,y_1,y_2,z} \quad & \tilde{f}(w,x,y_1,y_2,z) = r_1^{\mathsf{T}}x + s_2^{\mathsf{T}}w + x^{\mathsf{T}}CFy_1 + r_2^{\mathsf{T}}\begin{pmatrix} F & G \end{pmatrix}\begin{pmatrix} y_1 \\ y_2 \end{pmatrix} + s_2^{\mathsf{T}}z \\
\text{subject to} \quad & A_1 x + B_1 w = b \\
& A_2 \begin{pmatrix} F & G \end{pmatrix}\begin{pmatrix} y_1 \\ y_2 \end{pmatrix} + B_2 z = b_2 \\
& w, x, y_1, y_2, z \geq \mathbf{0}
\end{aligned}
\tag{21}
$$

Notice that the program is missing the element $x^\mathsf{T} CG y_2$, which would make its optimal solutions identical to the optimal solutions of Eq. (1). We describe a more practical approach to reducing the dimensionality in Appendix B. This approach is based on singular value decomposition and may be directly applied to any bilinear program. The following theorem quantifies the maximum error when using the compressed program.

**Theorem 18.** *Let $f^*$ and $\tilde{f}^*$ be optimal solutions of Eq. (1) and Eq. (21) respectively. Then:*

$$\epsilon = |f^* - \tilde{f}^*| \leq \sqrt{\lambda}.$$

*Moreover, this is the maximal linear dimensionality reduction possible with this error without considering the constraint structure.*

*Proof.* We first show that indeed the error is at most $\sqrt{\lambda}$ and that any linearly compressed problem with the given error has at least $f$ dimensions. Using a mapping that preserves the feasibility of both programs, the error is bounded by:

$$\epsilon \leq \left| f\left(w, x, \begin{pmatrix} F & G \end{pmatrix} \begin{pmatrix} y_1 \\ y_2 \end{pmatrix}, z\right) - \tilde{f}\left(w, x, y_1, \begin{pmatrix} y_2 \\ z \end{pmatrix}\right) \right| = \left| x^\mathsf{T} CG y_2 \right|.$$

Denote the feasible region of $y_2$ as $Y_2$. From the orthogonality of $[F, G]$, we have that $\|y_2\|_2 \leq 1$ as follows:

$$
\begin{aligned}
y &= \begin{pmatrix} F & G \end{pmatrix} \begin{pmatrix} y_1 \\ y_2 \end{pmatrix} \\
\begin{pmatrix} F^\mathsf{T} \\ G^\mathsf{T} \end{pmatrix} y &= \begin{pmatrix} y_1 \\ y_2 \end{pmatrix} \\
G^\mathsf{T} y &= y_2 \\
\|G^\mathsf{T} y\|_2 &= \|y_2\|_2
\end{aligned}
$$

Then we have:

$$
\begin{aligned}
\epsilon &\leq \max_{y_2 \in Y_2} \max_{x \in X} \left| x^\mathsf{T} CG y_2 \right| \leq \max_{y_2 \in Y_2} \|CG y_2\|_2 \\
&\leq \max_{y_2 \in Y_2} \sqrt{y_2^\mathsf{T} G^\mathsf{T} C^\mathsf{T} CG y_2} \leq \max_{y_2 \in Y_2} \sqrt{y_2^\mathsf{T} L y_2} \leq \sqrt{\lambda}
\end{aligned}
$$

The result follows from Cauchy-Schwartz inequality, the fact that $C^\mathsf{T} C$ is symmetric, and Assumption 17. The matrix $L$ denotes a diagonal matrix of eigenvalues corresponding to eigenvectors of $G$.

Now, let $H$ be an arbitrary matrix that satisfies the preceding error inequality for $G$. Clearly, $H \cap F = \emptyset$, otherwise $\exists y, \|y\|_2 = 1$, such that $\|CH y\|_2 > \epsilon$. Therefore, we have $|H| \leq n - |F| \leq |G|$, because $|H| + |F| = |Y|$. Here $|\cdot|$ denotes the number of columns of the matrix. □

Alternatively, the bound can be proved by replacing the equality $A_1 x + B_1 w = b_1$ by $\|x\|_2 = 1$. The bound can then be obtained by Lagrange necessary optimality conditions. In these bounds we use $L_2$-norm; an extension to a different norm is not straightforward. Note also that this dimensionality reduction technique ignores the constraint structure. When the constraints have some special structure, it might be possible to obtain an even tighter bound. As described in the next section, the dimensionality reduction technique generalizes the reduction that Becker et al. (2004) used *implicitly*.

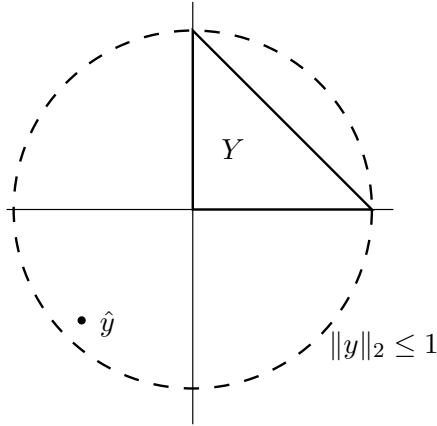

Figure 7: Approximation of the feasible set $Y$ according to Assumption 17.

The result of Theorem 18 is based on an approximation of the feasible set $Y$ by $\|y\|_2 \le 1$, as Assumption 17 states. This approximation may be quite loose in some problems, which may lead to a significant *multiplicative* overestimation of the bound in Theorem 18. For example, consider the feasible set depicted in Figure 7. The bound may be achieved in a point $\hat{y}$, which is far from the feasible region. In specific problems, a tighter bound could be obtained by either appropriately scaling the constraints, or using a weighted $L_2$ with a better precision. We partially address this issue by considering the structure of the constraints. To derive this, consider the following linear program and corresponding theorem:

$$\begin{aligned} \underset{x}{\text{maximize}} \quad & c^{\mathsf{T}}x \\ \text{subject to} \quad & Ax = b \quad x \ge \mathbf{0} \end{aligned} \tag{22}$$

**Theorem 19.** *The optimal solution of Eq. (22) is the same as when the objective function is modified to*

$$c^{\mathsf{T}}(I - A^{\mathsf{T}}(AA^{\mathsf{T}})^{-1}A)x,$$

*where $I$ is the identity matrix.*

*Proof.* The objective function is:

$$\begin{aligned} \max_{\{x \,|\, Ax=b,\, x \ge \mathbf{0}\}} c^{\mathsf{T}}x \quad &= \\ &= \max_{\{x \,|\, Ax=b,\, x \ge \mathbf{0}\}} c^{\mathsf{T}}(I - A^{\mathsf{T}}(AA^{\mathsf{T}})^{-1}A)x + c^{\mathsf{T}}A^{\mathsf{T}}(AA^{\mathsf{T}})^{-1}Ax \\ &= c^{\mathsf{T}}A^{\mathsf{T}}(AA^{\mathsf{T}})^{-1}b + \max_{\{x \,|\, Ax=b,\, x \ge \mathbf{0}\}} c^{\mathsf{T}}(I - A^{\mathsf{T}}(AA^{\mathsf{T}})^{-1}A)x. \end{aligned}$$

The first term may be ignored because it does not depend on the solution $x$. □

The following corollary shows how the above theorem can be used to strengthen the dimensionality reduction bound. For example, in zero-sum games, this stronger dimensionality reduction splits the bilinear program into two linear programs.

**Corollary 20.** *Assume that there are no variables $w$ and $z$ in Eq. (1). Let:*

$$\mathcal{Q}_i = (I - A_i^\mathsf{T}(A_i A_i^\mathsf{T})^{-1} A_i)), \quad i \in \{1, 2\},$$

*where $A_i$ are defined in Eq. (1). Let $\tilde{C}$ be:*

$$\tilde{C} = \mathcal{Q}_1 C \mathcal{Q}_2,$$

*where $C$ is the bilinear-term matrix from Eq. (1). Then the bilinear programs will have identical optimal solutions with either $C$ or $\tilde{C}$.*

*Proof.* Using Theorem 19, we can modify the original objective function in Eq. (1) to:

$$f(x, y) = r_1^\mathsf{T} x + x^\mathsf{T}(I - A_1^\mathsf{T}(A_1 A_1^\mathsf{T})^{-1} A_1) C (I - A_2^\mathsf{T}(A_2 A_2^\mathsf{T})^{-1} A_2)) y + r_2^\mathsf{T} y.$$

For the sake of simplicity we ignore the variables $w$ and $z$, which do not influence the bilinear term. Because both $(I - A_i^\mathsf{T}(A_i A_i^\mathsf{T})^{-1} A_i)$ for $i = 1, 2$ are orthogonal projection matrices, none of the eigenvalues in Theorem 18 will increase. $\square$

The dimensionality reduction presented in this section is related to the idea of compound events used in CSA. Allen, Petrik, and Zilberstein (2008a, 2008b) provide a detailed discussion of this issue.

## 5. Offline Bound

In this section we develop an approximation bound that depends only on the number of points for which $g(y)$ is evaluated and the structure of the problem. This kind of bound is useful in practice because it provides performance guarantees without actually solving the problem. In addition, the bound reveals which parameters of the problem influence the algorithm's performance. The bound is derived based on the maximal slope of $g(y)$ and the maximal distance among the points.

**Theorem 21.** *To achieve an approximation error of at most $\epsilon$, the number of points to be evaluated in a regular grid with $k$ points in every dimension must satisfy:*

$$k^n \geq \left( \frac{\|C\|_2 \sqrt{n}^n}{\epsilon} \right),$$

*where $n$ is the number of dimensions of $Y$.*

The theorem follows using basic algebraic manipulations from the following lemma.

**Lemma 22.** *Assume that for each $y_1 \in Y$ there exists $y_2 \in Y$ such that $\|y_1 - y_2\|_2 \leq \delta$ and $\tilde{g}(y_2) = g(y_2)$. Then the maximal approximation error is:*

$$\epsilon = \max_{y \in Y} g(y) - \tilde{g}(y) \leq \|C\|_2 \delta.$$

*Proof.* Let $y_1$ be a point where the maximal error is attained. This point is in $Y$, because this set is compact. Now, let $y_2$ be the closest point to $y_1$ in $L_2$ norm. Let $x_1$ and $x_2$ be the best responses for $y_1$ and $y_2$ respectively. From the definition of solution optimality we can derive:

$$
\begin{aligned}
r_1^\mathsf{T} x_1 + r_2^\mathsf{T} y_2 + x_1^\mathsf{T} C y_2 &\leq r_1^\mathsf{T} x_2 + r_2^\mathsf{T} y_2 + x_2^\mathsf{T} C y_2 \\
r_1^\mathsf{T} (x_1 - x_2) &\leq -(x_1 - x_2)^\mathsf{T} C y_2.
\end{aligned}
$$

The error now can be expressed, using the fact that $\|x_1 - x_2\|_2 \leq 1$, as:

$$
\begin{aligned}
\epsilon = r_1^\mathsf{T} x_1 + r_2^\mathsf{T} y_1 + x_1^\mathsf{T} C y_1 &- r_1^\mathsf{T} x_2 - r_2^\mathsf{T} y_1 - x_2^\mathsf{T} C y_1 \\
&= r_1^\mathsf{T}(x_1 - x_2) + (x_1 - x_2)^\mathsf{T} C y_1 \\
&\leq -(x_1 - x_2)^\mathsf{T} C y_2 + (x_1 - x_2)^\mathsf{T} C y_1 \\
&\leq (x_1 - x_2)^\mathsf{T} C (y_1 - y_2) \\
&\leq \|y_1 - y_2\|_2 \frac{(x_1 - x_2)^\mathsf{T}}{\|(x_1 - x_2)\|_2} C \frac{(y_1 - y_2)}{\|y_1 - y_2\|_2} \\
&\leq \|y_1 - y_2\|_2 \max_{\{x \,|\, \|x\|_2 \leq 1\}} \max_{\{y \,|\, \|y\|_2 \leq 1\}} x^\mathsf{T} C y \\
&\leq \delta \|C\|_2
\end{aligned}
$$

The above derivation follows from Assumption 17, and the bound reduces to the matrix norm using Cauchy-Schwartz inequality. $\square$

Not surprisingly, the bound is independent of the local rewards and transition structure of the agents. Thus it in fact shows that the complexity of achieving a fixed approximation with a fixed interaction structure is linear in the problem size. However, the bounds are still exponential in the dimensionality of the space. Notice also that the bound is additive.

## 6. Experimental Results

We now turn to an empirical analysis of the performance of the algorithm. For this purpose we use the *Mars rover* problem described earlier. We compared our algorithm with the original CSA and with a mixed integer linear program (MILP), derived for Eq. (1) as Petrik and Zilberstein (2007b) describe. Although Eq. (1) can also be modeled as a linear complementarity problem (LCP) (Murty, 1988; Cottle et al., 1992), we do not evaluate that option experimentally because LCPs are closely related to MILPs (Rosen, 1986). We expect these two formulations to exhibit similar performance. We also do not compare to any of the methods described by Horst and Tuy (1996) and Bennett and Mangasarian (1992) due to their very different nature and high complexity, and because some of these algorithms do not provide any optimality guarantees.

In our experiments, we applied the algorithm to randomly generated problem instances with the same parameters that Becker et al. (2003, 2004) used. Each problem instance includes 2 rovers and 6 sites. At each site, the rovers can decide to perform an experiment or to skip the site. Performing experiments takes some time, and all the experiments must be performed in 15 time units. The time required to perform an experiment is drawn from a discrete normal distribution with the mean uniformly chosen from 4.0-6.0. The variance

---

**Algorithm 5**: MPBP: Multiagent Planning with Bilinear Programming

---

**1** Formulate DEC-MDP $\mathcal{M}$ as a bilinear program $\mathcal{B}$ ;                    `// [Section 2.1]`
**2** $\mathcal{B}' \leftarrow$ ReduceDimensionality($\mathcal{B}$) with $\epsilon \leq 10^{-4}$ ;       `// [Section 4, Appendix B]`
**3** Convert $\mathcal{B}'$ to a semi-compact form ;                    `// [Definition 2]`
**4** $h \leftarrow -\infty$ ;
   `// Presolve step: run Algorithm 1 θ times with random initialization`
**5** **for** $i \in \{1 \ldots \theta\}$ **do**
**6**  $\quad\lfloor\ h \leftarrow \max\{h, \text{IterativeBestResponse}(\mathcal{B}')\}$ ;            `// [Algorithm 1]`
**7** BestResponseApprox($\mathcal{B}', \epsilon_0$) ;                    `// [Algorithm 2]`

---

is 0.4 of the mean. The local reward for performing an experiment is selected uniformly from the interval [0.1,1.0] for each site and it is identical for both rovers. The global reward, received when both rovers perform an experiment on a *shared site*, is super-additive and is 1/2 of the local reward. The experiments were performed with sites $\{1, 2, 3, 4, 5\}$ as shared sites. Typically, the performance of the algorithm degrades with the number of shared sites. Because the problem with fewer than 5 shared sites–as used in the original CSA paper–were too easy to solve, we only present results for problems with 5 shared sites. Note that CSA was used on this problem with an implicit dimensionality reduction due to the use of the *compound events*.

In these experiments, the naive dimensionality of $Y$ in Eq. (5) is $6 * 15 * 2 = 180$. This dimensionality can be reduced to be one per each shared site using the automatic dimensionality reduction procedure. Each dimension then represents the probability that an experiment on a *shared* site is performed regardless of the time. Therefore, the dimension represents the sum of the individual probabilities. Becker et al. (2004) achieved the same compression using *compound events*, where each compound event represents the fact that an experiment is performed on some site regardless of the specific time.

The complete algorithm–Multiagent Planning with Bilinear Programming (MPBP)–is summarized in Algorithm 5. The automatic dimensionality reduction reduces $Y$ to 5 dimensions. Then, reformulating the problem to a semi-compact form increases the dimensionality to 6. We experimented with different configurations of the algorithm that differ in the way the refinements of the pivot point selection is performed. The different methods, described in Section 3.3, were used to create six configurations as shown in Figure 8. The configuration $C_1$ corresponds to an earlier version of the algorithm (Petrik & Zilberstein, 2007a).

We executed the algorithm 20 times with each configuration on every problem, randomly generated according to the distribution described above. The results represent the average over the random instances. The maximum number of iterations of the algorithm was 200. Due to rounding errors, we considered any error less than $10^{-4}$ to be 0. The algorithm is implemented in MATLAB release 2007a. The linear solver we used is MOSEK version 5.0. The hardware configuration was Intel Core 2 Duo 1.6 GHz Low Voltage with 2GB RAM. The time to perform the dimensionality reduction is negligible and not included in the result.

A direct comparison with CSA was not possible because CSA cannot solve problems with this dimensionality within a reasonable amount of time. However, in a very similar problem setup with at most 4 shared sites, CSA solved only 76% of the problems, and the longest solution took approximately 4 hours (Becker et al., 2004). In contrast, MPBP solved all 200 problems with 4 shared sites *optimally* in less than 1 second on average, about 10000 times faster. In addition, MPBP returns solutions that are guaranteed to be close to optimal in the first few iterations. While CSA also returns solutions close to optimal very rapidly, it takes a very long time to confirm that.

| Configuration | Feasible [Eq. (13)] | Linear bound [Eq. (14)] | Cutting plane [Eq. (20)] | Presolve $[\theta]$ |
|---|---|---|---|---|
| $C_1$ | | | | 0 |
| $C_2$ | $\checkmark$ | | | 0 |
| $C_3$ | $\checkmark$ | $\checkmark$ | | 0 |
| $C_4$ | $\checkmark$ | | $\checkmark$ | 0 |
| $C_5$ | $\checkmark$ | $\checkmark$ | | 10 |
| $C_6$ | $\checkmark$ | | $\checkmark$ | 10 |

Figure 8: The six algorithm configurations that were evaluated. *Feasible, linear bound,* and *cutting plane* refer to methods used to determine the optimal solution.

Figure 9 shows the average guaranteed ratio of the optimal solution, achieved as a function of the number of iterations, that is, points for which $g(y)$ is evaluated. This figure, as all others, shows the result of the online error bound. This value is guaranteed and is *not* based on the optimal solution. This compares the performance of the various configurations of the algorithm, without using the presolve step. While the optimal solution was typically discovered in the first few iterations, it takes significantly longer to prove its optimality.

The average of absolute errors in both linear and log scale are shown in Figure 10. These results indicate that the methods proposed to eliminate the dominated region in searching for the pivot point can dramatically improve performance. While requiring that the new pivot points are feasible in $Y$ improves the performance, it is much more significant with a better approximation of $Y_h$. As expected, the *cutting plane* elimination is most efficient, but also most complex.

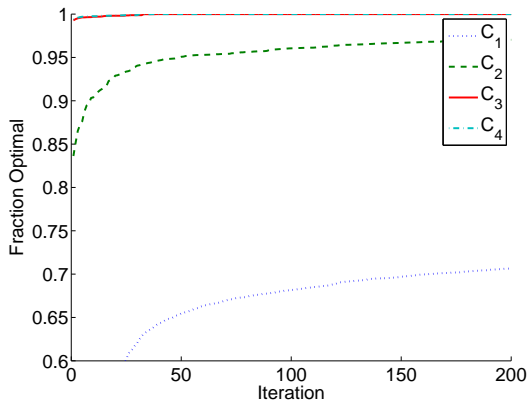

Figure 9: Guaranteed fraction of optimality according to the online bound.

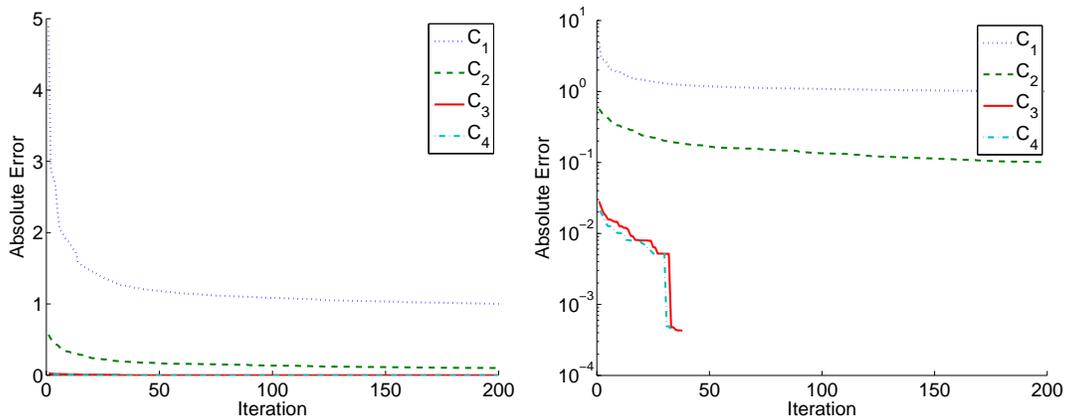

Figure 10: Comparison of absolute error of various region elimination methods.

To evaluate the tradeoffs in the implementation, we also show the average time per iteration and the average total time in Figure 11. These figures show that the time per iteration is significantly larger when the cutting plane elimination is used. Overall, the algorithm is faster when the simpler linear bound is used.

This trend is most likely problem specific. In problems with higher dimensionality, the more precise cutting plane algorithm may be more efficient. Implementation issues play a significant role in this problem too, and it is likely that the implementation of Algorithm 4 can be further improved.

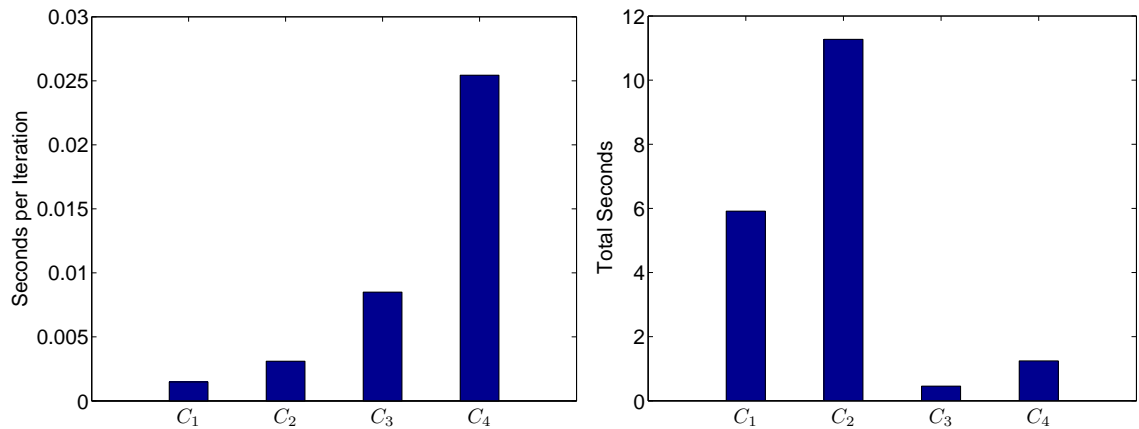

Figure 11: Time per iteration and the total time to solve. With configurations $C_1$ and $C_2$, the optimal value is not reached with 200 iterations . The figure only shows the time to compute up to 200 iterations.

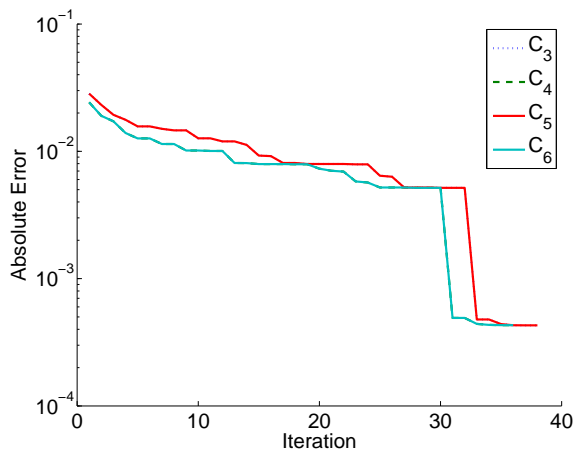

Figure 12: Influence of the presolve method.

Figure 12 shows the influence of using the presolve method. The plots of $C_3$ and $C_4$ are identical to the plots of $C_5$ and $C_6$ respectively, indicating that the presolve method does not have any significant influence. This also indicates that a solution that is very close to optimal is obtained when the values of the initial points are calculated.

We also performed experiments with CPLEX–a state-of-the-art MILP solver on the direct MILP formulation of the DEC-MDP. CPLEX was not able to solve any of the problems within 30 minutes, no matter how many of the sites were shared. The main reason for this is that it does not take any advantage of the limited interaction. Nevertheless, it is possible that some specialized MILP solvers may perform better.

## 7. Conclusion and Further Work

We present an algorithm that significantly improves the state-of-the-art in solving two-agent coordination problems. The algorithm takes as input a bilinear program representing the problem, and solves the problem using a new successive approximation method. It provides a useful online performance bound that can be used to decide when the approximation is good enough. The algorithm can take advantage of the limited interaction among the agents, which is translated into a small dimensionality of the bilinear program. Moreover, using our approach, it is possible to reduce the dimensionality of such problems automatically, without extensive modeling effort. This makes it easy to apply our new method in practice. When applied to DEC-MDPs, the algorithm is much faster than the existing CSA method, on average reducing computation time by four orders of magnitude. We also show that a variety of other coordination problems can be treated within this framework.

Besides multiagent coordination problems, bilinear programs have been previously used to solve problems in operations research and global optimization (Sherali & Shetty, 1980; White, 1992; Gabriel, Garca-Bertrand, Sahakij, & Conejo, 2005). Global optimization deals with finding the optimal solutions to problems with multi-extremal objective function. Solution techniques often share the same idea and are based on cutting plane methods. The main idea is to iteratively restrict the set of feasible solutions, while improving the incumbent solution. Horst and Tuy (1996) provide an excellent overview of these techniques. These algorithms have different characteristics and cannot be directly compared to the algorithm we developed. Unlike these traditional algorithms, we focus on providing quickly good approximate solutions with error bounds. In addition, we exploit the small dimensionality of the best-response space $Y$ to get tight approximation bounds.

Future work will address several interesting open questions with respect to the bilinear formulation as well as further improvement of the efficiency of the algorithm. With regard to the representation, it is yet to be determined whether the anytime behavior can be exploited when applied to games. That is, it is necessary to verify that an approximate solution to the bilinear program is also a meaningful approximation of the Nash equilibrium. It is also important to identify the classes of extensive games that can be efficiently formulated as bilinear programs.

The algorithm we present can be made more efficient in several ways. In particular, a significant speedup could be achieved by reducing the size of the individual linear programs. The programs are solved many times with the same constraints, but a different objective function. The objective function is always from a small-dimensional space. Therefore, the problems that are solved are all very similar. In the DEC-MDP domain, one option would be to use a procedure similar to action elimination. In addition, the performance could be significantly improved by starting with a tight initial triangulation. In our implementation, we simply use a single large polyhedron that covers the whole feasible region. A better approach would be to start with something that approximates the feasible region more tightly. A tighter approximation of the feasible region could also improve the precision of the dimensionality reduction procedure. Instead of the naive ellipsis used in Assumption 7, it is possible to use one that approximates the feasible region as tightly as possible. It is however very encouraging to see that even without these improvements, the algorithm is very effective compared with existing solution techniques.

## Acknowledgments

We thank Chris Amato, Raghav Aras, Alan Carlin, Hala Mostafa, and the anonymous reviewers for useful comments and suggestions. This work was supported in part by the Air Force Office of Scientific Research under Grants No. FA9550-05-1-0254 and FA9550-08-1-0181, and by the National Science Foundation under Grants No. IIS-0535061 and IIS-0812149.

## Appendix A. Proofs

**Proof of Proposition 10**  The proposition states that in the proposed triangulation, the sub-polyhedra do not overlap and they cover the whole feasible set $Y$, given that the pivot point is in the interior of $S$.

*Proof.* We prove the theorem by induction on the number of polyhedron splits that were performed. The base case is trivial: there is only a single polyhedron, which covers the whole feasible region.

For the inductive case, we show that for any polyhedron $S$ the sub-polyhedra induced by the pivot point $\hat{y}$ cover $S$ and do not overlap. The notation we use is the following: $T$

denotes the original polyhedron and $\hat{y} = Tc$ is the pivot point, where $\mathbf{1}^\mathsf{T}c = 1$ and $c \geq \mathbf{0}$. Note that $T$ is a matrix and $c, d, \hat{y}$ are vectors, and $\beta$ is a scalar.

We show that the sub-polyhedra cover the original polyhedron $S$ as follows. Take any $a = Td$ such that $\mathbf{1}^\mathsf{T}d = 1$ and $d \geq \mathbf{0}$. We show that there exists a sub-polyhedron that contains $a$ and has $\hat{y}$ as a vertex. First, let

$$\hat{T} = \begin{pmatrix} T \\ \mathbf{1}^\mathsf{T} \end{pmatrix}$$

This matrix is square and invertible, since the polyhedron is non-empty. To get a representation of $a$ that contains $\hat{y}$, we show that there is a vector $o$ such that for some $i$, $o(i) = 0$:

$$\begin{pmatrix} a \\ 1 \end{pmatrix} = \hat{T}d \;\; = \;\; \hat{T}o + (\beta \hat{y})$$
$$o \;\; \geq \;\; \mathbf{0},$$

for some $\beta > 0$. This will ensure that $a$ is in the sub-polyhedron with $\hat{y}$ with vertex $i$ replaced by $\hat{y}$. The value $o$ depends on $\beta$ as follows:

$$o = d - \beta \hat{T}^{-1} \begin{pmatrix} \hat{y} \\ 1 \end{pmatrix}.$$

This can be achieved by setting:

$$\beta = \min_i \frac{d(i)}{(\hat{T}^{-1}\hat{y})(i)}.$$

Since both $d$ and $c = \hat{T}^{-1}\hat{y}$ are non-negative. This leaves us with an equation for the sub-polyhedron containing the point $a$. Notice that the resulting polyhedron may be of a smaller dimension than $n$ when $o(j) = 0$ for some $i \neq j$.

To show that the polyhedra do not overlap, assume there exists a point $a$ that is common to the interior of at least two of the polyhedra. That is, assume that $a$ is a convex combination of the vertices:

$$a \;\; = \;\; T_3 c_1 + h_1 \hat{y} + \beta_1 y_1$$
$$a \;\; = \;\; T_3 c_2 + h_2 \hat{y} + \beta_2 y_2,$$

where $T_3$ represents the set of points common to the two polyhedra, and $y_1$ and $y_2$ represent the disjoint points in the two polyhedra. The values $h_1$, $h_2$, $\beta_1$, and $\beta_2$ are all scalars, while $c_1$ and $c_2$ are vectors. Notice that the sub-polyhedra differ by at most one vertex. The coefficients satisfy:

$$
\begin{array}{ll}
c_1 \geq \mathbf{0} & \qquad\qquad c_2 \geq \mathbf{0} \\
h_1 \geq 0 & \qquad\qquad h_2 \geq 0 \\
\beta_1 \geq 0 & \qquad\qquad \beta_2 \geq 0 \\
\mathbf{1}^\mathsf{T}c_1 + h_1 + \beta_1 = 1 & \qquad\qquad \mathbf{1}^\mathsf{T}c_2 + h_2 + \beta_2 = 1
\end{array}
$$

Since the interior of the polyhedron is non-empty, this convex combination is unique.

First assume that $h = h_1 = h_2$. Then we can show the following:

$$
\begin{aligned}
a = T_3 c_1 + h\hat{y} + \beta_1 y_1 &= T_3 c_2 + h\hat{y} + \beta_2 y_2 \\
T_3 c_1 + \beta_1 y_1 &= T_3 c_2 + \beta_2 y_2 \\
\beta_1 y_1 &= \beta_2 y_2 \\
\beta_1 = \beta_2 &= 0
\end{aligned}
$$

This holds since $y_1$ and $y_2$ are independent of $T_3$ when the polyhedron is nonempty and $y_1 \neq y_2$. The last equality follows from the fact that $y_1$ and $y_2$ are linearly independent. This is a contradiction, since $\beta_1 = \beta_2 = 0$ implies that the point $a$ is not in the interior of two polyhedra, but at their intersection.

Finally, assume WLOG that $h_1 > h_2$. Now let $\hat{y} = T_3 \hat{c} + \alpha_1 y_1 + \alpha_2 y_2$, for some scalars $\alpha_1 \geq 0$ and $\alpha_2 \geq 0$ that represent a convex combination. We get:

$$
\begin{aligned}
a = T_3 c_1 + h_1 \hat{y} + \beta_1 y_1 &= T_3(c_1 + h_1 \hat{c}) + (h_1 \alpha_1 + \beta_1) y_1 + h_1 \alpha_2 y_2 \\
a = T_3 c_2 + h_2 \hat{y} + \beta_2 y_2 &= T_3(c_2 + h_2 \hat{c}) + h_2 \alpha_1 y_1 + (h_2 \alpha_2 + \beta_2) y_2.
\end{aligned}
$$

The coefficients sum to one as shown below.

$$
\mathbf{1}^\mathsf{T}(c_1 + h_1 \hat{c}) + (h_1 \alpha_1 + \beta_1) + h_1 \alpha_2 = \mathbf{1}^\mathsf{T} c_1 + \beta_1 + h_1(\mathbf{1}^\mathsf{T}\hat{c} + \alpha_1 + \alpha_2) = \mathbf{1}^\mathsf{T} c_1 + \beta_1 + h_1 = 1
$$

$$
\mathbf{1}^\mathsf{T}(c_2 + h_2 \hat{c}) + \alpha_1 + (h_2 \alpha_2 + \beta_2) = \mathbf{1}^\mathsf{T} c_2 + \beta_2 + h_2(\mathbf{1}^\mathsf{T}\hat{c} + \alpha_1 + \alpha_2) = \mathbf{1}^\mathsf{T} c_2 + \beta_2 + h_2 = 1
$$

Now, the convex combination is unique, and therefore the coefficients associated with each vertex for the two representations of $a$ must be identical. In particular, equating the coefficients for $y_1$ and $y_2$ results in the following:

$$
\begin{aligned}
h_1 \alpha_1 + \beta_1 &= h_2 \alpha_1 & h_1 \alpha_2 &= h_2 \alpha_2 + \beta_2 \\
\beta_1 &= h_2 \alpha_1 - h_1 \alpha_1 & \beta_2 &= h_1 \alpha_2 - h_2 \alpha_2 \\
\beta_1 &= \alpha_1(h_2 - h_1) > 0 & \beta_2 &= \alpha_2(h_1 - h_2) < 0
\end{aligned}
$$

We have that $\alpha_1 > 0$ and $\alpha_2 > 0$ from the fact that $\hat{y}$ is in the interior of the polyhedron $S$. Then, having $\beta_2 \leq 0$ is a contradiction with $a$ being a convex combination of the vertices of $S$. $\qquad \square$

## Appendix B. Practical Dimensionality Reduction

In this section we describe an approach to dimensionality reduction that is easy to implement. Note that there are at least two possible approaches to take advantage of reduced dimensionality. First, it is possible to use the dimensionality information to limit the algorithm to work only in the significant dimensions of $Y$. Second, it is possible to modify the bilinear program to have a small dimensionality. While changing the algorithm may be more straightforward, it limits the use of the advanced pivot point selection methods described in Section 3.3. Here, we show how to implement the second option in a straightforward way using singular value decomposition.

The dimensionality reduction is applied to the following bilinear program:

$$\begin{aligned}
\underset{w,x,y,z}{\text{maximize}} \quad & r_1^{\mathsf{T}} x + s_1^{\mathsf{T}} w + x^{\mathsf{T}} C y + r_2^{\mathsf{T}} y + s_2^{\mathsf{T}} z \\
\text{subject to} \quad & A_1 x + B_1 w = b_1 \\
& A_2 y + B_2 z = b_2 \\
& w,x,y,z \geq \mathbf{0}
\end{aligned} \tag{23}$$

Let $C = SVT^{\mathsf{T}}$ be a singular value decomposition. Let $T = [T_1, T_2]$, such that the singular value of vectors $t_i$ in $T_2$ is less than the required $\epsilon$. Then, a bilinear program with reduced dimensionality may be defined as follows:

$$\begin{aligned}
\underset{w,x,\bar{y},y,z}{\text{maximize}} \quad & r_1^{\mathsf{T}} x + s_1^{\mathsf{T}} w + x^{\mathsf{T}} S V T_1 \bar{y} + r_2^{\mathsf{T}} y + s_2^{\mathsf{T}} z \\
\text{subject to} \quad & T_1 \bar{y} = y \\
& A_1 x + B_1 w = b_1 \\
& A_2 y + B_2 z = b_2 \\
& w,x,y,z \geq \mathbf{0}
\end{aligned} \tag{24}$$

Note that $\bar{y}$ is not constrained to be non-negative. One problematic aspect of reducing the dimensionality is how to define the initial polyhedron that needs to encompass all feasible solutions. One option is to make it large enough to contain the set $\{y \mid \|y\|_2 = 1\}$, but this may be too large. Often in practice, it may be more efficient to first triangulate a rough approximation of the feasible region, and then execute the algorithm on this triangulation.

## Appendix C. Sum of Convex and Concave Functions

In this section we show that the best-response function $g(y)$ may not be convex when the program is not in a semi-compact form. The convexity of the best-response function is crucial in bounding the approximation error and in eliminating the dominated regions.

We show that when the program is not in a semi-compact form, the best-response function can we written as a sum of a convex function and a concave function. To show that consider the following bilinear program.

$$\begin{aligned}
\underset{w,x,y,z}{\text{maximize}} \quad & f = r_1^{\mathsf{T}} x + s_1^{\mathsf{T}} w + x^{\mathsf{T}} C y + r_2^{\mathsf{T}} y + s_2^{\mathsf{T}} z \\
\text{subject to} \quad & A_1 x + B_1 w = b_1 \\
& A_2 y + B_2 z = b_2 \\
& w,x,y,z \geq \mathbf{0}
\end{aligned} \tag{25}$$

This problem may be reformulated as:

$$\begin{aligned}
f &= \max_{\{y,z \mid (y,z) \in Y\}} \max_{\{x,w \mid (x,w) \in X\}} r_1^{\mathsf{T}} x + s_1^{\mathsf{T}} w + x^{\mathsf{T}} C y + r_2^{\mathsf{T}} y + s_2^{\mathsf{T}} z \\
&= \max_{\{y,z \mid (y,z) \in Y\}} g'(y) + s_2^{\mathsf{T}} z,
\end{aligned}$$

where

$$g'(y) = \max_{\{x,w \,|\, (x,w) \in X\}} r_1^\mathsf{T} x + s_1^\mathsf{T} w + x^\mathsf{T} C y + r_2^\mathsf{T} y.$$

Notice that function $g'(y)$ is convex, because it is a maximum of a set of linear functions. Since $f = \max_{\{y \,|\, (y,z) \in Y\}} g(y)$, the best-response function $g(y)$ can be expressed as:

$$\begin{aligned} g(y) &= \max_{\{z \,|\, (y,z) \in Y\}} g'(y) + s_2^\mathsf{T} z = g'(y) + \max_{\{z \,|\, (y,z) \in Y\}} s_2^\mathsf{T} z \\ &= g'(y) + t(y), \end{aligned}$$

where

$$t(y) = \max_{\{z \,|\, A_2 y + B_2 z = b_2, \; y,z \geq \mathbf{0}\}} s_2^\mathsf{T} z.$$

Function $g'(y)$ does not depend on $z$, and therefore could be taken out of the maximization. The function $t(y)$ corresponds to a linear program, and its dual using the variable $q$ is:

$$\begin{aligned} \underset{q}{\text{minimize}} \quad & (b_2 - A_2 y)^\mathsf{T} q \\ \text{subject to} \quad & B_2^\mathsf{T} q \geq s_2 \end{aligned} \tag{26}$$

Therefore:

$$t(y) = \min_{\{q \,|\, B_2^\mathsf{T} q \geq s_2\}} (b_2 - A_2 y)^\mathsf{T} q,$$

which is a concave function, because it is a minimum of a set of linear functions. The best-response function can now be written as:

$$g(y) = g'(y) + t(y),$$

which is a sum of a convex function and a concave function, also known as a d.c. function (Horst & Tuy, 1996). Using this property, it is easy to construct a program such that $g(y)$ will be convex on one part of $Y$ and concave on another part of $Y$, as the following example shows. Note that in semi-compact bilinear programs $t(y) = 0$, which guarantees the convexity of $g(y)$.

**Example 23.** *Consider the following bilinear program:*

$$\begin{aligned} \underset{x,y,z}{\text{maximize}} \quad & -x + xy - 2z \\ \text{subject to} \quad & -1 \leq x \leq 1 \\ & y - z \leq 2 \\ & z \geq 0 \end{aligned} \tag{27}$$

*A plot of the best response function for this program is shown in Figure 13.*

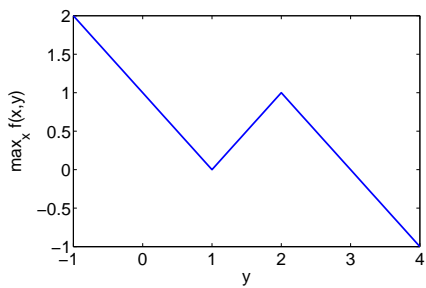

Figure 13: A plot of a non-convex best-response function $g$ for a bilinear program, which is not in a semi-compact form.